\documentclass{article}

\usepackage[table]{xcolor}
\usepackage{hyperref}
\usepackage{url}
\usepackage{graphicx}
\usepackage{booktabs} 
\usepackage{array}
\usepackage{wrapfig}
\usepackage{enumerate}
\usepackage{subcaption}
\usepackage{multirow}
\usepackage{floatrow}
\usepackage{dblfloatfix}
\usepackage{tablefootnote}
\usepackage{amsmath,amssymb} 
\usepackage{adjustbox}
\usepackage{subcaption}
\usepackage{makecell}   
\usepackage{paralist}
\newcommand{\coloredurl}[2][blue]{%
  \href{#2}{\textcolor{#1}{\nolinkurl{#2}}}%
}
\usepackage{tabularx}   % flexible column widths
\newcolumntype{L}[1]{>{\raggedright\arraybackslash}p{#1}} % fixed-width left
\newcolumntype{C}[1]{>{\centering\arraybackslash}p{#1}}   % fixed-width center
\newcolumntype{X}{>{\raggedright\arraybackslash}X}         % auto width
\usepackage{pifont}     % \ding symbols for ✓ and ✗
\newcommand{\xmark}{\ding{55}}  % x‑mark
\usepackage{xcolor}
\usepackage[most]{tcolorbox}
\usepackage{placeins}
\tcbset{
  inline highlight/.style={
    on line,
    boxsep=1pt,             % inner padding
    left=2pt,right=2pt,     % horizontal padding
    top=1pt,bottom=1pt,     % vertical padding
    arc=2pt,                % slightly rounded corners
    colback=green!30,       % fill colour
    colframe=green!30,      % frame colour = fill
    fontupper=\footnotesize,% size inside the box
    halign=center           % center the text
  }
}

\usepackage[accepted]{icml2025}

\usepackage{mathtools}
\usepackage{amsthm}

\usepackage[capitalize,noabbrev]{cleveref}

\theoremstyle{plain}

\theoremstyle{definition}

\theoremstyle{remark}

\usepackage[textsize=tiny]{todonotes}

\icmltitlerunning{Using Multiple Input Modalities Can Improve Data-Efficiency and O.O.D. Generalization for ML with Satellite Imagery}

\begin{document}

\twocolumn[
\icmltitle{Using Multiple Input Modalities Can Improve Data-Efficiency and O.O.D. Generalization for ML with Satellite Imagery}

\icmlsetsymbol{equal}{*}

\begin{icmlauthorlist}
\icmlauthor{Arjun Rao}{cu}
\icmlauthor{Esther Rolf}{cu}

\end{icmlauthorlist}

\icmlaffiliation{cu}{Department of Computer Science, University of Colorado Boulder}

\icmlcorrespondingauthor{Arjun Rao}{raoarjun@colorado.edu}

\icmlkeywords{Machine Learning for Earth Observation, Label Efficiency, OOD generalization}

\vskip 0.3in
]
\printAffiliationsAndNotice{}

\begin{abstract}
A large variety of geospatial data layers is available around the world ranging from remotely-sensed raster data like satellite imagery, digital elevation models, predicted land cover maps, and human-annotated data, to data derived from environmental sensors such as air temperature or wind speed data. A large majority of machine learning models trained on satellite imagery \textbf{(SatML)}, however, are designed primarily for \emph{optical} input modalities such as multi-spectral satellite imagery. To better understand the value of using other input modalities alongside optical imagery in supervised learning settings, we generate augmented versions of SatML benchmark tasks by appending additional geographic data layers to datasets spanning classification, regression, and segmentation. Using these augmented datasets, we find that fusing additional geographic inputs with optical imagery can significantly improve SatML model performance. Benefits are largest in settings where labeled data are limited and in geographic out-of-sample settings, suggesting that multi-modal inputs may be especially valuable for data-efficiency and out-of-sample performance of SatML models. Surprisingly, we find that  hard-coded fusion strategies outperform learned variants, with interesting implications for future work.
\end{abstract}

\section{Introduction}
SatML models that effectively leverage the volume and diversity of data from Earth Observation (EO) satellites have the potential to translate petabyte-scale raw data into data-driven insights. Users of SatML systems need models that can integrate these vast arrays of publicly available geographic data into a cohesive representation of the world, allowing for accurate predictions even with limited training data, or when faced with covariate shifts across time, space, spectrum, and scale \citep{rolfposition}. 
%
% Multi-modal, geographic inputs to SatML models have yet to be evaluated in these mission-critical evaluation settings. 

While including additional input layers is clearly likely to increase performance for in-sample prediction with ample training data, the effects of adding additional input layers in settings with \emph{limited label data} and \emph{out-of-sample deployment} distributions are less clear. Additional geographic inputs could inform a SatML model with structural information that may allow the model to learn geospatial image representations with fewer labeled training samples (label-efficiency); they could also require more complex (data-hungry) models to represent the various modalities of data. Additional inputs could help SatML models generalize across regions; they could also cause models to overfit to local patterns that only manifest in-sample, which could then decrease performance. 

\textbf{In this work, we study the label-efficiency and out-of-sample generalization capability associated with adding non-optical, contextual inputs to commonly used SatML architectures.}  

As outlined in \citet{datacentric}, data-centric learning is a systematic method of algorithmic evaluation where the primary focus involves curating diverse, complete, unbiased, and relevant data for optimal model performance. We perform a \emph{data-centric} study on the benefits and nuances of leveraging these widely available geographic input layers, complementing previous lines of model-centric research that study how to utilize multi-modal inputs for a fixed training/pretraining strategy and/or model architecture. 
% Across a wide variety of input datasets, real-world task-types, and model architectures, we observe performance improvements when these geospatial inputs are fused with traditional optical imagery in critical label-scarce and out-of-distribution test settings. Ablation experiments indicate that arbitrary fusion/learning strategies may not necessarily help SatML model performance in critical label-scarce and out-of-distribution deployment settings, presenting challenges and opportunities that inform future work in multi-modal ML for EO.

Our primary findings in this work are:
\begin{inparaenum}[(1)]
  \item We show improvements in label-efficiency when multi-modal, auxiliary geographic inputs are fused with optical imagery on 3 SatML task-types: Multi-label land-cover classification, land cover segmentation, and tree-cover regression.
  \item We find that these auxiliary geographic inputs are especially helpful when SatML models are evaluated OOD through results on the spatially buffered test split of the BigEarthNetv2.0 dataset \citep{bigearthnetv2}, and the OOD test cities of the EnviroAtlas dataset in Austin, TX, and Durham, NC \citep{ipm}. 
  \item Through our ablations, we find surprising results that show the ineffectiveness of finetuning SatML models arbitrarily on common benchmark tasks with these auxiliary geographic inputs. 
\end{inparaenum}

Our contributions also include a large-scale, multi-dataset release containing modified versions of the SustainBench farmland boundary delineation dataset \citep{sustainbench}, and the USAVars tree-cover regression dataset \citep{mosaiks} with additional geographic inputs georeferenced to the optical imagery. Additionally, we release the BigEarthNetv2.0 dataset \citep{bigearthnetv2} with pre-computed patch-embeddings with the SatCLIP location encoder \citep{satclip}. A full list of contributed data products is shown in column ``Additional Data Layers'' in \Cref{tab:dataset_modalities}.

% \begin{itemize}
%     \item We propose an experimental and evaluation framework that aims to understand \emph{how} and \emph{when} auxiliary geographic inputs can improve SatML model performance in the critical label-scarce and out-of-sample settings. A component of this contribution includes extensive benchmarking on 4 SatML benchmark datasets spanning 3 commonly observed task-types.  
%     \item We release modified versions of the EnviroAtlas land cover segmentation dataset \citep{enviroatlas}, SustainBench farmland boundary delineation dataset \citep{sustainbench}, and the USAVars tree-cover regression dataset \citep{mosaiks} with additional geographic inputs georeferenced to the optical imagery. A full list of contributed data products is shown in column "XX" \Cref{tab:dataset_modalities}.
%     % \item Our ablation experiments in \Cref{sec:ablations} propose important modeling guidelines that cautions practitioners on the dangers of utilizing these auxiliary inputs arbitrarily.
% \end{itemize}

\section{Prior Work}
\subsection{Multi-Modal SatML}
\label{sec:prior_work} Adding a non-optical context to machine learning models trained on geospatial imagery has been performed extensively in prior work. \citet{loc1} extracts GPS features from the Yahoo Flickr Creative Commons 100M dataset, and fuses embeddings of location information with final embeddings from a convolution-based image network. \citet{loc2} incorporates geolocation information into fine-grained image classification through the use of geolocation priors, introducing the computer vision community to geo-aware neural networks. \citet{mac2019presence} performed fine-grained image classification with a location, time, and photographer prior to differentiate between similar classes that are spatially disparate. \citet{benson2024multi} add a contextual input to predict future vegetation state given temporally rich satellite imagery and future weather information. \citet{wang2020urban2vec} propose an unsupervised multi-modal framework which incorporates both street view imagery and point-of-interest data to learn neighborhood embeddings in urban areas. \citet{opensentinelmap, osm2, osm3} introduce large-scale Sentinel-2 datasets georeferenced with OpenStreetMap (OSM) rasters \citep{osm} converted to be used as a land-use-land-cover map (LULC). However, these methods, which utilize geographic data layers publicly available, intend for their usage to be restricted as ground-truth masks for land-cover classification problems. 

Recently, \citet{mmearth} introduce large, multi-modal pre-training datasets built with Sentinel-2 imagery that contain several geographic modalities like ESA WorldCover \citep{zanaga2022esa} and Digital Elevation Model. Although MMEarth \citep{mmearth} is pre-trained on these modalities, it is only used to predict the modalities given a Sentinel-2 RGB image as input; nonetheless, they find data-efficiency improvements when their self-supervised models are linear-probed on various downstream classification tasks. \citet{multimae} utilize the Aster-DEM and the ESA-Worldcover raster produced by \citet{mmearth} as additional input to a masked autoencoder (MAE). However, a bulk of their experiments is performed with various permutations of Sentinel-2-derived multispectral modalities.

\subsection{Token Fusion}
Studies on Vision Transformers (ViTs) have explored the use of additional tokens to improve performance and capture more nuanced information. In \citet{vit}, a \emph{class token} (\texttt{[CLS]}) was introduced and appended to the patch embeddings, enabling the model to learn a global representation useful for classification tasks. \citet{deit} introduce a \emph{distillation token} to facilitate knowledge transfer from a teacher model, boosting accuracy without substantially increasing computational cost. \citet{vpt, vpt2} demonstrate that injecting a small set of learnable prompts into the early layers of pre-trained ViTs can effectively adapt them to new downstream tasks. \citet{registers} highlights the importance of internal ``registers'' in ViT architectures, arguing that specialized design choices can better accommodate these additional tokens for more robust representations. 

% Beyond supervised settings, self-supervised approaches such as \cite{ssltransformer} employ additional tokens in a teacher-student framework to capture rich semantic features through contrastive objectives. The fusion of these tokens typically relies on concatenation with the existing patch tokens and identical processing in Transformer layers, though some works adopt cross-attention or gating mechanisms for selective token interaction. Notably, this practice is reminiscent of strategies in natural language processing (NLP), where models like BERT \citep{bert} rely on specialized tokens (e.g., \texttt{[CLS]}, \texttt{[SEP]}) to encode global representations and manage sequence boundaries. These developments highlight the adaptability of token-based designs in Transformers and underscore their potential to enrich task-specific representations across both vision and language domains.

\section{Methods}
Our experiments measure performance of models trained with just multi-spectral inputs and with additional geographic inputs. We consider three different fusion mechanisms that allow for SatML models to learn from these geographic auxiliary inputs. We then describe the model architectures used for each fusion mechanism, and the benchmark datasets that we train our models on.
%
% \textbf{Design Choices: Model architectures and geographic data layers: }Our design-choices of models to utilize for each benchmark dataset is informed by availability of prior benchmark results. For example, we use a U-Net for field-boundary delineation on the SustainBench dataset due to prior established results in \citet{ermonsus}, a ViT for BigEarthNet multi-label classification similar to \citet{bigearthnetv2}, and a 5-layer FCN on the EnviroAtlas land-cover segmentation dataset to be consistent with results reported in \citet{ipm}. We choose geographic data-layers that we believe are intuitively helpful to the downstream task. 

\begin{figure*}[t!]
 \centering
 \includegraphics[width=\linewidth]{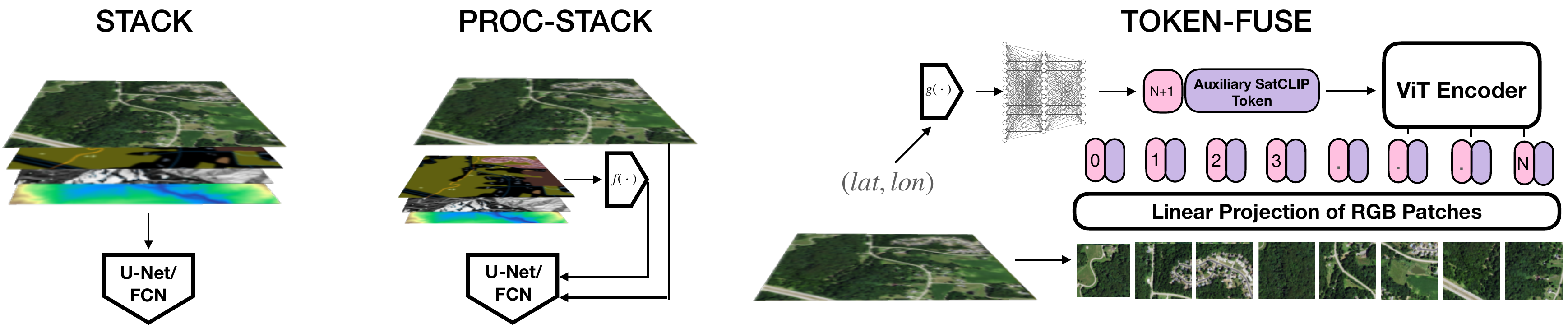}
 \caption{\textbf{Geographic data input fusion mechanisms used in this work: } \texttt{STACK} involves concatenating one or more  geographic raster inputs with the optical input before passed jointly as an input to a convolution-based architecture. \texttt{PROC-STACK} passes the geographic input to a function $f(\cdot)$ before stacking the geographic data with the optical input. \texttt{TOKEN-FUSE} passes a latitude-longitude pair to a location encoder $g(\cdot)$ and uses location embeddings as an auxiliary token to a Vision Transformer (ViT). 
 %Most of our  perform ablations with a fine-tuned auxiliary token.
 Experiments in \Cref{sec:results-data-efficiency} and \Cref{sec:results-OOD} use frozen models for $f$ and $g$; ablation experiments in \Cref{sec:ablations} use trainable models.
 }
 \label{fig:schematic}
\end{figure*}   

\subsection{Geographic Data Fusion}
  
\Cref{fig:schematic} contains an overview of the proposed geographic input-fusion techniques used in this work. For land cover segmentation with the EnviroAtlas dataset, we fuse the original inputs (NAIP aerial imagery) with roads, waterways, and waterbody data from the OSM repository \citep{osm} using the fusion method \texttt{STACK}. We compute the hand-crafted prior for the training split in Pittsburgh, and test splits in Austin and Durham using the methodology proposed in \citet{ipm}. The generation of the prior is denoted by $f(\cdot)$ in \Cref{fig:schematic}, and is described in detail in appendix \Cref{sec:priorgen}. The resulting prior along with the raw geographic data layers are used as input to the prior function and are fused to the SatML. The generation of the prior followed by fusion with the optical input forms our fusion method \texttt{PROC-STACK}.

For the farmland-parcel delineation task with the SustainBench dataset, and the socioeconomic regression task with the USAVars dataset, we use OSM raster layers that contain all the geographic data layers used for the EnviroAtlas dataset, with the addition of several new land-use and land-cover classes that are roughly relevant to the task. These additional raster layers include high-level biome information such as forests, wetlands, or urban-type terrain. Output Geodataframes are pre-processed to RGB space. We apply a smoothing kernel ($\sigma = 1.0$) to remove sharp edges and features from the API response. A complete list of raster inputs queried for the USAVars dataset is detailed in appendix \Cref{fig:sample_osm}. Additionally, we pull a digital elevation map (DEM) from the Continental Europe Digital Terrain Model available as part of the OpenTopography API. The DEM raster, originally available at a $20$m GSD, is resized to the Sentinel-2 RGB spatial resolution of $10$m/px. Unlike the OSM rasters, the DEM is passed as raw input with fusion mechanism \texttt{STACK}.

To be comparable to previous benchmark results, we use a fully convolutional network (FCN) for the EnviroAtlas \citet{ipm} Dataset, a U-Net \citep{unet} for the SustainBench-field-delineation dataset \citet{sustainbench}, and a ResNet50 \citet{resnet} for the regression task proposed in the USAVars dataset \citet{mosaiks}. 

For the BigEarthNetv2.0 image-level multi-label classification task we use vision transformer (ViT, ViT-B/8, ViT-S/8) architectures. To the Sentinel-2 input, we fuse general-purpose global SatCLIP location embeddings \citep{satclip}, which distill socioeconomic and environmental signals in satellite imagery into a pretrained location encoder $g(\textrm{lat,lon})$ with output dimension 256. Embeddings from SatCLIP's location encoder are passed as an auxiliary token to the ViT's encoder along with image tokens. We add a linear layer to SatCLIP's location encoder that maps the 256-dimensional SatCLIP embeddings to the desired sequence length expected by the Vit-S/ViT-B. The auxiliary SatCLIP token is assigned a positional encoding of $N+1$ where $N$ is the total number of encoder tokens excluding the classification token. For our main experiments, the parameters within the SatCLIP model $g(\textrm{lat,lon})$ are frozen; we experiment with unfreezing these weights in \Cref{fig:ft_satclip} and \Cref{sec:ablations}.

\begin{table*}[htb]
\footnotesize
\caption{\textbf{Experimental framework and source tasks used in this work:} We test fusion mechanisms \texttt{STACK} and \texttt{STACK-PROC} on the EnviroAtlas \citep{ipm}, SustainBench \citep{sustainbench}, and the USAVars \citep{mosaiks} benchmark datasets. We test fusion mechanism \texttt{TOKEN-FUSE} on the BigEarthNetv2.0 \citep{bigearthnetv2} classification dataset. Labels queried that form OSM rasters are shown in appendix \Cref{fig:sample_osm}. \textdagger~denotes geographic data layers released with this work (aligned with the benchmark datasets). }
\label{tab:dataset_modalities}
\begin{tabularx}{\textwidth}{ L{2.5cm} X L{2.6cm} c L{3.2cm} C{1.0cm}@{} }
\toprule
\textbf{Dataset} & \textbf{Task Description} & \textbf{Multispectral Input} & \textbf{Model} & \textbf{Additional Data Layers} & \textbf{OOD?} \\[0.1em]
\midrule
SustainBench \citep{sustainbench} & Farmland boundary delineation & Sentinel-2 RGB & U-Net &
  OSM rasters\textdagger, EU-DEM\textdagger & \xmark \\[0.3em]

EnviroAtlas \citep{ipm} & Land-cover segmentation & NAIP RGB + NIR & FCN &
  Prior \citep{ipm}, OSM rasters & \checkmark \\[0.1em]

BigEarthNetv2.0 \citep{bigearthnetv2} & Land-cover classification & Sentinel-2 (10 bands) & ViT &
  SatCLIP \citep{satclip} embeddings\textdagger & \checkmark \\[0.5em]

USAVars \citep{mosaiks} & Tree-cover regression & NAIP RGB + NIR & ResNet-50 &
  OSM rasters\textdagger & \xmark \\[0.3em]
\bottomrule
\end{tabularx}
\end{table*}

\subsection{Models}
\textbf{Convolutional Architectures: } 
 In this work, we use simple, widely-used convolutional neural networks when trained on data fused with fusion mechanisms \texttt{STACK} and \texttt{PROC-STACK}. We choose simple architectures over specialized SatML model architectures because we are primarily interested in comparing different data settings and fusion strategies. We choose models to be consistent with model architectures used in prior work. For experiments on the EnviroAtlas dataset, we use a 5-layer FCN. For segmentation on the SustainBench field-boundary delineation, we use a U-Net \citep{unet} with identical architectural setup and hyperparameters as \citet{ermonsus} to allow for consistency when comparing results. For regression on the USAVars tree-cover dataset, we use a vanilla ResNet50 \citep{resnet} with randomly initialized weights.

\textbf{Vision Transformers (ViTs): } Vision Transformers (ViTs) \citep{vit} utilize the transformer architecture proposed in \citep{vaswani2017attention}. Input images are decomposed into a sequence of small, non-overlapping patches which are mapped to embeddings (tokens) with a linear-layer projection.  
% \verb|[CLS]| is a learnable additional token introduced to capture label information. 
Unlike \citep{satmae, scalemae} that use various versions of sinusoidal positional encodings that are sensitive to Ground Sampling Distance (GSD) and temporal information, we augment image patches with learnable positional encodings. 

\textbf{Learned location encoders: } Location encoders in SatML help models interpolate to new geographic regions by incorporating terrain and environmental signals given a (lat, lon) pair. SatCLIP \citep{satclip} builds on GeoCLIP \citep{geoclip}, CSP \citep{CSP}, and GPS2Vec \citep{gps2vec} by integrating a CLIP-inspired \citep{clip} contrastive learning framework specifically designed for satellite imagery from the Sentinel-2 EO satellite. SatCLIP's location encoder, which can be used out-of-the-box, accurately captures terrain, environmental, and socioeconomic signals \citep{satclip}. Unlike the previously used convolutional architectures that accept a rasterized input of geographic data projected to the correct Coordinate Reference System (CRS), models trained with the SatCLIP location encoder accept embeddings as an auxiliary token.

\subsection{Datasets}
We conduct experiments using 4 benchmark datasets in ML for remote sensing. These datasets cover different prediction tasks, multi-spectral input sources, and additional data layers used. \Cref{tab:dataset_modalities} presents an overview of the datasets and additional layers used. 
\emph{All additional geographic data layers, georeferenced with benchmark datasets, are available as a hosted dataset at }\coloredurl[magenta]{https://huggingface.co/datasets/arjunrao2000/geolayers}. We release our code that allows for training models on our datasets at \coloredurl[magenta]{https://github.com/arjunarao619/geolayers-terrabytes}.

\textbf{BigEarthNet (Classification): }The BigEarthNetv2.0 dataset \citep{bigearthnet, bigearthnetv2} is a multi-label classification task that consists of approximately 550,000 pairs of Sentinel-2 image patches, paired with ground labels of over 19 land cover classes. Our models input 10 Sentinel-2 bands to ensure consistency with benchmark results reported in \citet{bigearthnetv2}. Unlike the original BigEarthNet dataset in \citet{bigearthnet}, BigEarthNetv2.0 \citet{bigearthnetv2} constructs a training, validation, and test split by using a grid-based split assignment algorithm. Validation and test areas-of-sampling are not within the geographic extent of the training area-of-sampling, ensuring no data-leakage. Thus, our results reported on the BigEarthNetv2.0 dataset can be considered an out-of-sample validation and test.

\textbf{EnviroAtlas (Land Cover Segmentation): } The EnviroAtlas dataset (compiled by \citet{ipm} and composed of data from \citet{enviroatlas}) consists of high-resolution ($1$m) land cover maps derived from NAIP aerial imagery. In this dataset, coarse land-cover maps from the National Land Cover Database (NLCD) are aligned with buildings, road networks, water bodies, and waterways from public sources such as the OSM project \citep{osm}. 
The ``prior'' data layer constructed in \citet{ipm} is a (hand-coded) fusion of NLCD data with OSM data, in the form of \texttt{PROC-STACK}. EnviroAtlas's train split only covers the Pittsburgh region. We use the provided out-of-sample validation and test datasets in Austin and Durham and in-distribution validation and test datasets in Pittsburgh. 

\textbf{SustainBench (Field Boundary Delineation) } The SustainBench benchmark proposed in \citet{sustainbench} contains a collection of 15 benchmark tasks in machine learning for remote sensing spanning 7 United Nations' sustainable development goals (SDGs). We use the field-delineation task which consists of Sentinel-2 imagery in France in 2017. Each input image is at a $10$m ground-sampling distance and has a size of $224 \times 224$ pixels corresponding to an approximately $5$ km$^2$ surface area covered per image. 

\textbf{USAVars Tree-cover (Regression): } \\
The USAVars dataset proposed in \citet{mosaiks} comprises approximately 100,000 pairs of NAIP aerial imagery cropped to a spatial extent of $~$1-sq-km per image containing real-valued labels of tree-cover, population density. We pull rasters of several land-cover and infrastructure-related classes from OSM \citep{osm} as a geographic input, aligned to the RGB layers. Our final set of labels cover broad biome-related land-cover classes such as waterbodies, forests, and buildings with fine-grained labels covering sub-categories of biomes. A complete list of labels pulled from OSM are shown in appendix \Cref{fig:sample_osm}. \\

\begin{figure}[t]
  \centering
  \includegraphics[width=\linewidth]{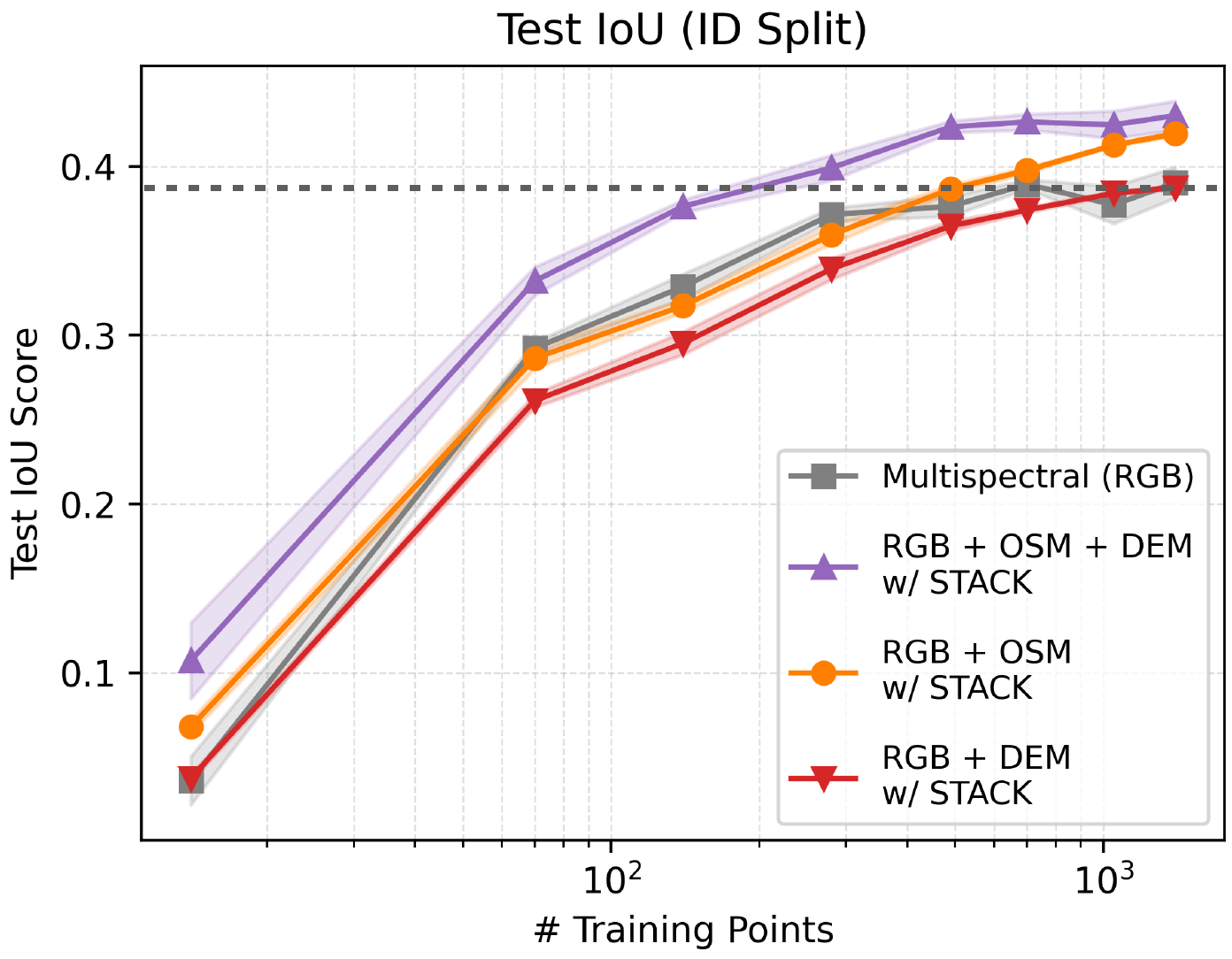}
  \caption{\textbf{Performance and label-efficiency of a U-Net trained on SustainBench's Farmland Boundary Delineation Dataset.} We use the standard ID split as benchmarked on in \cite{ermonsus}. Label efficiency and out-of-distribution performance reported as IoU scores averaged over five random seeds. OSM and EU-DEM-aided models match RGB-only model’s best score with 221 training images (total = 1573 images).}
  \label{fig:sustainbench}
\end{figure}

\begin{figure}[h!]
  \centering
  \includegraphics[width=\linewidth]{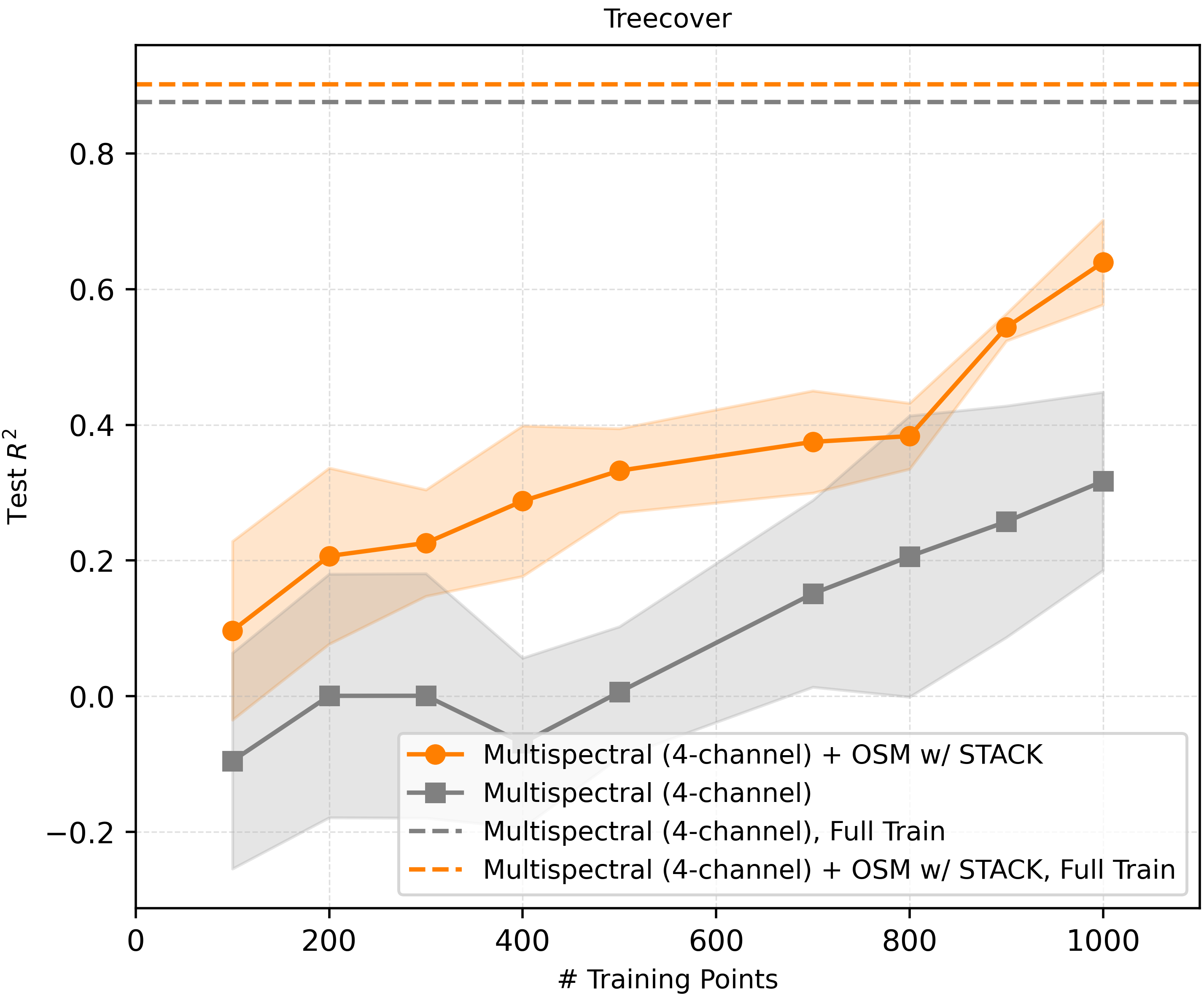}
  \caption{\textbf{Performance and data-efficiency of a ResNet50 trained with and without an OSM raster data layer as \texttt{STACK} on the USAVars treecover regression task.} Dashed lines show the performance of each input set using the full dataset (100,000 points) as training data.
  }
  \label{fig:usavars_treecover}
\end{figure}

\section{Results} \label{sec:results}

Across all four SatML benchmark datasets covering tasks in semantic segmentation and multi-label classification  for land cover, field boundary delineation, and regression, we found that adding contextual, geographic inputs improves model performance, with largest gains in settings with limited label data (\Cref{sec:results-data-efficiency}) and out-of-distribution test sets (\Cref{sec:results-OOD}). Ablation experiments (\Cref{sec:ablations} ) provide evidence that fine-tuning encoders aided by geographic input layers does not necessarily help in these critical settings.

\begin{table*}[t]
    \centering
    \small
    \caption{\textbf{Comparison of a ViT's Average Precision and 
      multi-label F1 score (Macro-averaged) on the BigEarthNetv2.0 test split with and 
      without a SatCLIP location encoder auxiliary token.} BigEarthNetv2.0 \citep{bigearthnet} consists of $549,488$ Sentinel-2 image tiles. Ablation includes linear probing a pre-trained SatCLIP location encoder (Right). Mean results over five random seeds. Unless specified, all results report $\leq 0.1\%$ standard error.}
    \label{tab:vit_table}
    \vspace{0.5em}
    \begin{tabular}{
        l
        cccc
        cccc
        >{\centering\arraybackslash}p{1.4cm}
    }
    \toprule
    \textbf{Subset (\%)} 
      & \multicolumn{4}{c}{\textbf{ViT-B}} 
      & \multicolumn{4}{c}{\textbf{ViT-S}} 
      & \textbf{SatCLIP} \\
    \cmidrule(lr){2-5} \cmidrule(lr){6-9}
      & \multicolumn{2}{c}{\textbf{W/ SatCLIP Aux.\ Token}} 
      & \multicolumn{2}{c}{\textbf{Vanilla ViT}}
      & \multicolumn{2}{c}{\textbf{W/ SatCLIP Aux.\ Token}} 
      & \multicolumn{2}{c}{\textbf{Vanilla ViT}}
      & \\
    \cmidrule(lr){2-3} \cmidrule(lr){4-5} \cmidrule(lr){6-7} \cmidrule(lr){8-9}
      & Avg Prec     & F1            & Avg Prec      & F1
      & Avg Prec     & F1            & Avg Prec      & F1
      & F1 \\
    \midrule
    1\%   & \textbf{46.3} & \textbf{36.1} & 44.6          & 32.1
          & 40.45         & 23.27~$\pm$~1.27 & 39.78      & 22.95~$\pm$~1.31
          & 15.9 \\
    2\%   & \textbf{55.6} & \textbf{45.9} & 51.1          & 40.2
          & 47.96         & 33.82~$\pm$~1.10 & 45.60      & 34.11
          & 14.1 \\
    5\%   & \textbf{62.7} & \textbf{54.1} & 58.9          & 50.2
          & 59.98         & 47.84~$\pm$~2.08 & 56.07      & 44.05~$\pm$~1.13
          & 10.1 \\
    20\%  & \textbf{66.8} & \textbf{60.6} & 64.5          & 58.1
          & 66.4          & 58.3             & 64.2       & 57.6
          & 12.5 \\
    50\%  & \textbf{70.1} & \textbf{64.7} & 68.7          & 63.5
          & 70.1          & 64.3             & 69.2       & 63.7
          & 21.7 \\
    100\% & 70.3          & 65.2          & 69.5          & 64.1
          & \textbf{70.8} & \textbf{65.4}    & 70.1       & 64.5
          & 23.2 \\
    \bottomrule
    \end{tabular}
\end{table*}

\subsection{Geographic inputs can aid data-efficiency}
\label{sec:results-data-efficiency}

The benefit of additional geographic data inputs on data-efficiency of SatML models can be seen in all four experimental settings and all three fusion mechanisms.

From \Cref{fig:sustainbench}, we see performance improvements with low amounts of training data when using the \texttt{STACK} approach to fuse additional raster layers. A U-Net trained with an OSM and DEM raster layer using fusion mechanism \texttt{STACK} exhibits an $8.1\%$ test dice score improvement in-sample when trained on between 1-5$\%$ of training data on the SustainBench field-boundary delineation dataset, compared to a $4.1\%$ improvement when using the full training dataset. From appendix \Cref{tab:susbench_model_ablations}, we find that these performance improvements hold with most commonly used SatML segmentation model architectures introduced over the past five years. From \Cref{fig:usavars_treecover}, stacking OSM raster layers as input to a ResNet-50 for the USAVars tree-cover regression task improves $R^2$ by $0.162$ points when trained on between 60 to 250 training images. This performance improvement reduces to a 0.026 improvement in $R^2$ when the full 68,000 image training dataset is used. 

From \Cref{tab:enviroatlas}, we find that a prior generated and fused with \texttt{PROC-STACK} improves in-distribution test accuracy of land-cover segmentation on the EnviroAtlas dataset \citep{enviroatlas} by $9.3\%$ when trained on between $1$ to $5\%$ of the training dataset, compared to a $0.6\%$ improvement when trained with the full training dataset. When the raw data-layers used to generate the prior in \citet{ipm} are fused with fusion mechanism \texttt{STACK} before training, data-efficiency improvements drop to approximately $2\%$ over ten random seeds for this range ($1-5\%$) of training data, still an improvement.

\begin{figure*}[ht]
  \centering
  \includegraphics[width=\textwidth]{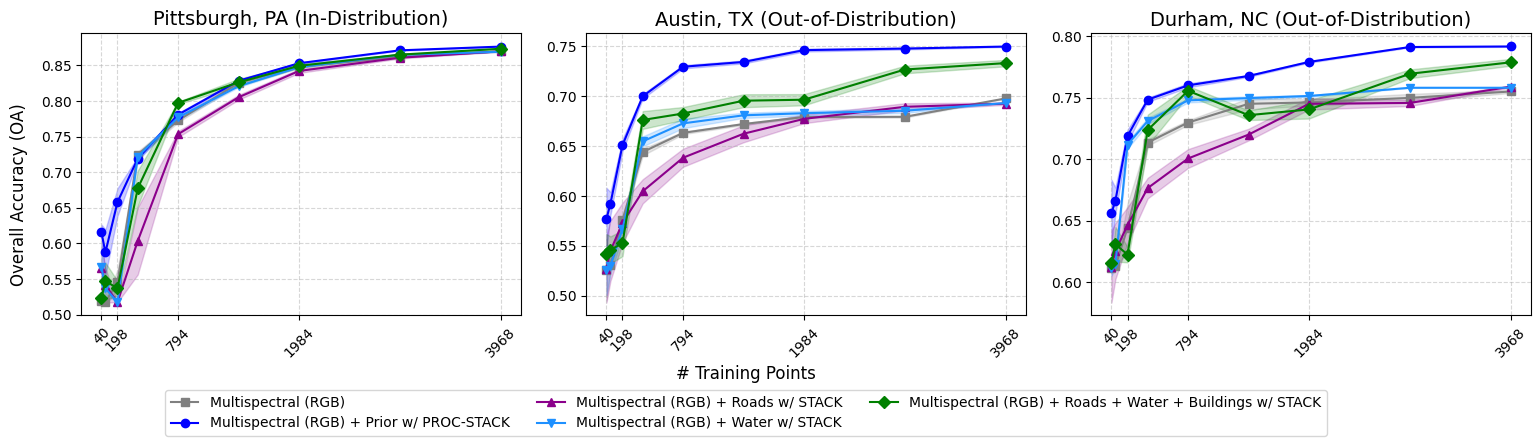}
  \caption{\textbf{Performance and label-efficiency of a FCN on the EnviroAtlas Land Cover Segmentation Dataset with \texttt{STACK} and \texttt{PROC-STACK} geographic input fusion}. Austin and Durham are out-of-sample test splits. Results averaged over $10$ random seeds. $1 \times$ standard error of Pittsburgh reported $\leq 1e^{-3}$ over 10 random seeds. 
  %Label-efficiency results shown in \Cref{tab:enviroatlas}
  }
  \label{fig:enviroatlas}
\end{figure*}
\begin{table*}[h!]
  \centering
  \footnotesize
  \begin{tabular*}{\textwidth}{@{\extracolsep{\fill}}l c
    >{\columncolor{gray!20}}c >{\columncolor{gray!20}}c
    c >{\columncolor{gray!20}}c >{\columncolor{gray!20}}c
    c >{\columncolor{gray!20}}c >{\columncolor{gray!20}}c@{}}
    \toprule
    \textbf{Subset (\%)} &
      \multicolumn{3}{c}{\textbf{Pittsburgh}} &
      \multicolumn{3}{c}{\textbf{Austin}} &
      \multicolumn{3}{c}{\textbf{Durham}} \\
    \cmidrule(lr){2-4}\cmidrule(lr){5-7}\cmidrule(lr){8-10}
        & \textbf{RGB} & \textbf{Prior} & \textbf{All}
        & \textbf{RGB} & \textbf{Prior} & \textbf{All}
        & \textbf{RGB} & \textbf{Prior} & \textbf{All}\\
    \midrule
    1\% & 0.51 & \textbf{0.61} & 0.52
         & 0.53 $\pm$ 0.03 & \textbf{0.58 $\pm$ 0.03} & 0.54 $\pm$ 0.02
         & 0.61 $\pm$ 0.00 & \textbf{0.66 $\pm$ 0.03} & 0.62 $\pm$ 0.00\\
    2\% & 0.51 & \textbf{0.58} & 0.54
         & 0.53 $\pm$ 0.01 & \textbf{0.59 $\pm$ 0.01} & 0.55 $\pm$ 0.01
         & 0.61 $\pm$ 0.00 & \textbf{0.67 $\pm$ 0.01} & 0.63 $\pm$ 0.01\\
    5\% & 0.54 & \textbf{0.65} & 0.55
         & 0.58 $\pm$ 0.00 & \textbf{0.65 $\pm$ 0.01} & 0.55 $\pm$ 0.01
         & 0.64 $\pm$ 0.02 & \textbf{0.72 $\pm$ 0.01} & 0.62 $\pm$ 0.01\\
    \bottomrule
  \end{tabular*}
  \caption{\textbf{Performance of EnviroAtlas prior, all raster inputs versus RGB input with $1\%, 2\%$, and $5\%$ of input training data.}}
  \label{tab:enviroatlas}
\end{table*}

On the SustainBench field-boundary delineation and the USAVars tree-cover regression datasets, we note that largest gains in label-efficiency are observed with training dataset sizes of 100-700 images, which we observe to be the low-data-regime where geographic input layers consistently outperform models trained on optical modalities. For example, on the USAVars tree-cover regression task, we observe a diminished gap in the test $R^2$ metric as we scale from 700 training samples ($\Delta_{R^{2}} = 0.36$) to 1400 training samples ($\Delta_{R^{2}} = 0.08$).

We also note that not \emph{all} geographic inputs/combinations of these inputs improve label-efficiency and OOD performance when fused with the SatML model using the fusion mechanisms introduced in \Cref{fig:schematic}. In \Cref{fig:enviroatlas}, we note that a road-map raster worsens performance compared to standard, multispectral-only training. Similarly, from \Cref{fig:sustainbench}, concatenating a single DEM raster to optical imagery for a field-boundary delineation task on the SustainBench dataset hurts performance in these settings.

\begin{table}[h!]
\centering
\small
\begin{tabular}{cccccc}
\toprule
Sub\% & \multicolumn{2}{c}{\texttt{PROC-STACK}, FCN$_{\text{out}}$} & RGB Only & \texttt{STACK} \\
\cmidrule(r){2-3}
      & 1 & 3 &  & \\
\midrule
1\%   & \textbf{40.8/26.3} & 35.7/22.7  & 5.5/2.9 & 26.5/10.1 \\
5\%   & \textbf{49.5/33.7} & 47.0/31.5  & 45.7/29.9 & 47.0/31.6 \\
10\%  & 52.7/36.6 & 53.0/36.9  & 49.0/32.6 & \textbf{54.7/37.9} \\
20\%  & 55.2/38.9  & 54.7/38.5 & 53.8/37.7 & \textbf{57.3/39.9} \\
35\%  & 56.8/40.5  & 55.9/39.7 & 54.6/38.4 & \textbf{59.3/42.5} \\
50\%  & 57.1/40.9 & 57.0/40.7  & 56.3/38.7 & \textbf{60.3/42.8} \\
75\%  & 58.1/41.8 & 58.5/42.3  & 56.9/40.3 & \textbf{60.7/43.9} \\
100\% & 59.9/43.6 & 59.3/43.1 & 57.9/39.5 & \textbf{61.4/42.9} \\
\bottomrule
\end{tabular}
\caption{\textbf{Test Dice/IoU score when OSM and EU-DEM rasters are fused via a trainable FCN with \texttt{PROC-STACK} on the SustainBench field-boundary delineation task.} We allow the intermediate FCN to output (FCN$_{\text{out}}$) $1$ and $3$-channel raster outputs. Averaged over $5$ random seeds. $100\%$ corresponds to $1572$ total training points. RGB, \texttt{STACK} reported from \Cref{fig:sustainbench}.} 
\label{tab:sustainbench_ablation}
\end{table}
\begin{table}[h!]
\centering
\begin{tabular}{cccc}
\toprule
Sub\% & \texttt{F} SatCLIP & Register Token & \texttt{FT} SatCLIP \\
\midrule
1\%   & \textbf{46.3}/\textbf{36.1}  & 45.1/33.2 & 45.4/34.7 \\
2\%   & \textbf{55.6}/\textbf{45.9}  & 50.3/40.5 & 53.2/42.8 \\
5\%   & 62.7/54.1                  & 61.6/53.9 & \textbf{63.5}/\textbf{56.2} \\
20\%  & \textbf{66.8}/\textbf{60.6}  & 65.3/59.8 & 65.3/59.1 \\
50\%  & \textbf{70.1}/\textbf{64.7}  & 68.1/60.9 & 67.1/60.1 \\
100\% & \textbf{70.3}/\textbf{65.2}  & 66.5/59.6 & 66.0/59.1 \\
\bottomrule
\end{tabular}
\caption{\textbf{Average Precision (Macro)/ Multi-Label F1 score with Frozen (\texttt{F}) vs Register vs Fine-tuned (\texttt{FT}) SatCLIP auxiliary token on the BigEarthNetv2.0 dataset}. Results with a register token are reported with the addition of one register token to a ViT-B. 100\% corresponds to $\approx 430,000$ image patches. Results averaged over 5 random seeds.}
\label{tab:satclip_ablation}
\end{table}

\subsection{Geographic inputs can aid out-of-distribution performance}
\label{sec:results-OOD}
We also found that fusing additional geographic input layers to remotely sensed imagery can significantly aid geographic domain generalization. While the value of additional input layers is clear in low-label settings (here $<$800 training points) for all test cities in the EnviroAtlas dataset, \Cref{fig:enviroatlas} also shows an improvement in overall test accuracy across all amounts of training data for the out-of-distribution test cities in different states (Austin, TX and Durham, NC). We observe a $4.12\%$ improvement in the overall accuracy with the prior geographic data layer using \texttt{PROC-STACK} and a $2.03\%$ improvement when the raw raster data layers used to generate the prior are fused with \texttt{STACK}. Unlike the ID test set (Pittsburgh), the gains in performance in the OOD settings do not appear to diminish with more training samples, as the OOD performance curves remain significantly separated across settings, even using 100\% of the training data.

From \Cref{tab:vit_table}, performance improvements on the BigEarthNetv2.0 dataset with the auxiliary SatCLIP token fused with \texttt{TOKEN-FUSE} also hold over all training data subsets. This reflects OOD performance as the BigEarthNetv2.0  validation and test splits use a spatial buffering approach \cite{bigearthnetv2}.
For a ViT-B, we observe a $3.1\%$ improvement in the multi-label F1 metric, and a $2.5\%$ improvement in the multi-label average precision metric. Interestingly, for a ViT-S, this improvement in out-of-sample accuracy across all data subsets drops to a $2\%$ improvement in average precision and a $1\%$ improvement in the multi-label F1 metric. We hypothesize that this difference in performance can possibly be attributed to the reduced \emph{model expressivity} of ViT-S that prevents it from fully exploiting the SatCLIP auxiliary token (embedding size of $384$ vs $768$).

\subsection{Finetuning geographic-input aided SatML models can hurt label-efficiency and OOD performance} \label{sec:ablations}
To determine if geographic inputs that are learned during training aid label efficiency and out-of-sample generalization of SatML models on commonly used benchmark datasets, we conduct ablation studies for the fusion mechanisms \texttt{TOKEN-FUSE} and \texttt{PROC-STACK}. In sections \Cref{sec:results-data-efficiency,sec:results-OOD}, we freeze the intermediate modules $f(\cdot)$ in \texttt{PROC-STACK} and $g(\cdot)$ in \texttt{TOKEN-FUSE} ($f(\cdot)$ and $g(\cdot)$ from \Cref{fig:schematic}). In \Cref{sec:learned-compression,sec:finetuned-satclip}, we finetune these modules jointly with the SatML model.

\subsubsection{Learned compression with \texttt{PROC-STACK}}
\label{sec:learned-compression}
To understand when a compressed embedding of geographic rasters can confer similar results as using all as input, we design a trainable \texttt{PROC-STACK} fusion mechanism used to train a U-Net on the SustainBench field boundary delineation task. In this approach, we pass both the DEM (1 channel) and OSM (19 channels) geographic data layers to a trainable FCN architecture. Outputs from the FCN are stacked with the original optical input and passed to the U-Net, and both models are trained simultaneously\footnote{To accommodate for the increased number of trainable parameters, we increase the number of epochs the models are trained on and allow for convergence.}.

Label efficiency on the SustainBench field boundary delineation dataset is shown in \Cref{tab:sustainbench_ablation}. The fusion mechanism \texttt{PROC-STACK} on learned, compressed inputs is not competitive with a simple \texttt{STACK} of the pre-processed, original rasters. Interestingly, we observe significantly improved label efficiency of the trained \texttt{PROC-STACK} ablation model between subsets $1\%$ and $5\%$. These label efficiency improvements, however, do not hold across all subsets.

\begin{figure}[t!]
    \centering
    \includegraphics[width=0.99\linewidth]{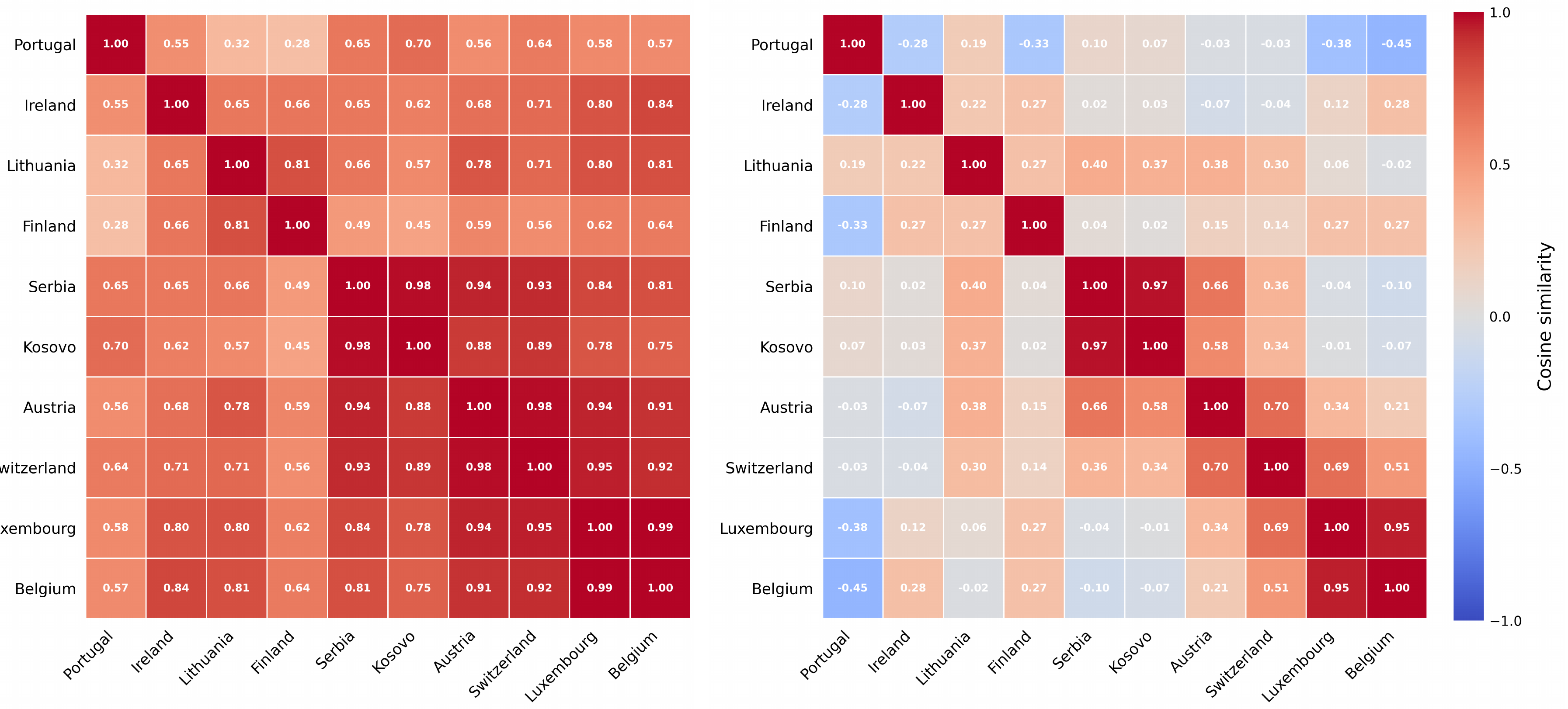}
    \caption{\textbf{Pairwise cosine similarity of SatCLIP embeddings used to form auxiliary ViT token: Frozen (Left) vs Fine-Tuned (Right). } On the BigEarthNetv2.0 land cover classification task, the fine-tuned SatCLIP token exhibits significantly greater pairwise disagreement between countries compared to the frozen token in countries covered by the train-split of BigEarthNetv2.0.}
    \label{fig:pairwise_sims}
\end{figure}

\subsubsection{Fine-tuning location encoders in \texttt{TOKEN-FUSE}}
\label{sec:finetuned-satclip}
For classification on the BigEarthNetv2.0 dataset, instead of using a frozen SatCLIP encoder with a learnable linear projection layer (as in \Cref{sec:results-OOD}), we now allow for the SatCLIP model to be trainable given the original pre-trained SatCLIP location encoder weights. 
%This novel training setting is equivalent to training the SatCLIP location encoder on an image-classification pseudo task. 

We find that the label-efficiency and out-of-sample performance degrade when the SatCLIP weights are learnable during training (\Cref{tab:satclip_ablation}). \Cref{fig:pairwise_sims} shows that fine-tuning the SatCLIP model in this fashion leads to embeddings that are highly localized within various countries covered by the BigEarthNetv2.0 dataset. This suggests that the augmented ViT may be overfitting to the auxiliary SatCLIP token, leading to lower test set performance when the SatCLIP model is trainable. Furthermore, overfitting is particularly likely considering that the trainable weights of the SatCLIP location encoder span $360$k parameters -- significantly higher than other image tokens input to the ViT. \footnote{Addition of layer-normalization to the SatCLIP token doesn't significantly alter performance, label-efficiency, and OOD generalization.}

To understand the performance discrepancy between the fine-tuned and frozen location encoders in the \texttt{TOKEN-FUSE} strategy,  we compare performance of our auxiliary SatCLIP token against a generic (non-geospatial) learnable register token as a baseline. First introduced in \cite{registers}, register tokens are randomly initialized, fully-trainable prefix tokens. Register tokens capture high-norm ``outlier" artifacts that hold significantly lower local-patch information. ViTs aided with registers show improvements only when trained with sufficiently large numbers of trainable parameters (ViT-B, ViT-L, ViT-H) over long training durations. We choose a ViT-B ($86$M trainable parameters) with identical hyperparameters as experiments that produced results in \Cref{tab:vit_table}.

From \Cref{tab:satclip_ablation}, we find that registers do not improve label efficiency and out-of-sample performance of a ViT-B trained on the BigEarthNetv2.0 dataset compared to a frozen SatCLIP location encoder. We find that adding additional register tokens up to 3 tokens doesn't significantly alter this result. Interestingly, both a register token and a fine-tuned SatCLIP token outperform a vanilla ViT-B when trained on between $1\%$ to $20\%$ of training data, but perform worse than a vanilla ViT in the large-data ($50\%$, $100\%$) regime.

% \begin{table}[t]
% \centering
% \small
% \begin{tabular}{cccc}
% \toprule
% Sub\% & \texttt{PROC-STACK} & RGB Only & \texttt{STACK} \\
% \midrule
% 1\%   & \textbf{40.8/26.3} & 5.5/2.9   & 26.5/10.1 \\
% 5\%   & \textbf{49.5/33.7} & 45.7/29.9 & 47.0/31.6 \\
% 10\%  & 52.7/36.6          & 49.0/32.6 & \textbf{54.7/37.9} \\
% 20\%  & 55.2/38.9          & 53.8/37.7 & \textbf{57.3/39.9} \\
% 35\%  & 56.8/40.5          & 54.6/38.4 & \textbf{59.3/42.5} \\
% 50\%  & 57.1/40.9          & 56.3/38.7 & \textbf{60.3/42.8} \\
% 75\%  & 58.1/41.8          & 56.9/40.3 & \textbf{60.7/43.9} \\
% 100\% & 59.9/43.6          & 57.9/39.5 & \textbf{61.4/42.9} \\
% \bottomrule
% \end{tabular}
% \caption{\textbf{Test Dice/IoU score when OSM and EU‑DEM rasters are fused via a trainable FCN with \texttt{PROC-STACK}.} We allow the intermediate FCN to output a single‐channel raster embedding. Averaged over five random seeds. RGB, \texttt{STACK} reported from \Cref{fig:sustainbench}.}
% \label{tab:sustainbench_ablation}
% \end{table}

\section{Experimental Takeaways}
\textbf{Takeaway 1: Auxiliary geographic inputs improve performance in low-data settings.}
In \Cref{sec:results-data-efficiency}, we find notable performance improvements in low-data settings with an auxiliary OSM and DEM geographic input layer (0.08 IoU on SustainBench, 9.3\% OA on EnviroAtlas, 0.162 R$^2$ improvement on USAVars). On the SustainBench field boundary delineation task, a U-Net trained with an OSM and EU-DEM raster matches the test IoU of an RGB-only model with only 224 training samples (compared to 1573 training samples for the RGB-only model).

\textbf{Takeaway 2: Auxiliary geographic inputs improve performance OOD.} From \Cref{sec:results-OOD}, we find that these geographic layers are especially helpful when evaluated on OOD splits of the benchmark datasets: 4.12\% improvement in EnviroAtlas's OOD cities, 3.1\% improvement on BigEarthNetv2.0's spatially-buffered test splits. 

\textbf{Takeaway 3: Finetuning SatML models aided by auxiliary geographic inputs \emph{can} hurt performance.} Surprisingly, when we allow the intermediate module in \texttt{PROC-STACK} (denoted by $f(\cdot)$ in \Cref{fig:schematic}) to be trainable and act as a geographic input compression module, test IoU scores drop, on average, by $4.1\%$ on the \Cref{tab:sustainbench_ablation}) test set. Higher performance drops occur as the expressivity of the intermediate FCN is increased from 1 to 3 output channels. From \Cref{sec:finetuned-satclip}, we find that allowing a SatCLIP encoder to be jointly trained with the SatML classification model causes the model to overfit (\Cref{fig:pairwise_sims}), hurting label efficiency and OOD performance in the BigEarthNet task.

% \section{Discussion}
% talk about roads
% \paragraph{Discussion: } 
% Improvements in label-efficiency and out-of-distribution SatML performance directly translate to applicability for real-world, downstream applications in climate and ecological monitoring. Here, we found that fusing additional geographic input layers into SatML models resulted in better label-efficiency and out-of-sample performance, compared to models that only used multispectral input. 

% To get these improvements, we used simple and effective multi-modal geospatial data layer fusion methodologies -- input stacking for raster layers, and adding auxiliary multi-modal tokens to a ViT for location embeddings. Recent work \cite{tile2vec, rotate, rapid, dw, seco} have shown the feasibility and efficiency of more advanced convolution-based neural network architectures that exploit the unique properties of geospatial imagery. We argue that the proposed fusion mechanisms and geographic inputs can be integrated into these models with ease, and with minimal modifications to the source architecture. Geographic data fusion can also be extended beyond the supervised setting. For example, reconstructing RGB inputs with the geographic input fused with \texttt{STACK} using a Masked Autoencoder should intuitively allow for better geospatial self-supervised models. For ViTs, we present the first result that highlights the benefit of \emph{geographic token-fusion}. 

\textbf{Limitations and future work: } In \Cref{fig:enviroatlas,fig:sustainbench,fig:vit_ben_plots,fig:usavars_treecover,tab:enviroatlas,tab:vit_table}, we use geographic data-layers that make sense for the downstream task. As we are primarily interested in potential benefits of using additional data layers, we restrict the scope of the study only to these geographic input layers and do not train on a larger corpus of raster and scalar inputs. Here, we use fusion mechanisms \texttt{STACK}, \texttt{PROC-STACK} for convolutional models and \texttt{TOKEN-FUSE} for ViTs since they involve minimal modifications to the source architectures; future work will examine more sophisticated fusion mechanisms.

\section*{Acknowledgements}
A majority of training runs conducted in this work were run on an NVIDIA Grace-Hopper (GH200) GPU node provided by the University of Colorado Boulder's high performance computing system Alpine. We thank Brandon Reyes and the RC computing team at CU Boulder for allowing access to this resource. Alpine is jointly funded by the University of Colorado Boulder, the University of Colorado Anschutz, Colorado State University, and the National Science Foundation (award 2201538). 

OpenStreetMap is open data, licensed under the \href{https://opendatacommons.org/licenses/odbl/}{Open Data Commons Open Database License} by the \href{https://osmfoundation.org/}{OpenStreetMap Foundation} (OSMF).

DEM data in this work is derived from services provided by the OpenTopography Facility with support from the National Science Foundation under NSF Award Numbers 2410799, 2410800 \& 2410801 \citep{opentopo}.

We thank Dr. Caleb Robinson for invaluable feedback during the writing stage of this work. We also thank the anonymous reviewers for their comments and suggestions.

\section*{Impact Statement}
By lowering annotation costs and delivering consistent accuracy when models cross regional, temporal or sensor boundaries, our approach can democratize high‑impact Earth‑observation applications such as crop monitoring, disaster assessment and biodiversity mapping for organizations with limited resources. Because the fusion layers are lightweight and the best results come from \emph{frozen} tokens using \emph{pretrained} encoders, our work avoids the large training footprints typical of foundation‑model fine‑tuning, mitigating energy use relative to existing alternatives. However, the work also surfaces risks: uneven coverage or quality in auxiliary datasets (e.g., OSM) could entrench geographic biases, and fine‑tuning the location encoder can cause severe overfitting to local patterns. Practitioners should therefore audit input‑layer availability and monitor model generalization before deployment in safety‑ or equity‑critical settings.

\bibliography{Geolayers_TerraBytes_main}

\begin{thebibliography}{47}
\providecommand{\natexlab}[1]{#1}
\providecommand{\url}[1]{\texttt{#1}}
\expandafter\ifx\csname urlstyle\endcsname\relax
  \providecommand{\doi}[1]{doi: #1}\else
  \providecommand{\doi}{doi: \begingroup \urlstyle{rm}\Url}\fi

\bibitem[Aung et~al.(2020)Aung, Uzkent, Burke, Lobell, and Ermon]{ermonsus}
Aung, H.~L., Uzkent, B., Burke, M., Lobell, D., and Ermon, S.
\newblock {F}arm {P}arcel {D}elineation using {S}patio-{T}emporal {C}onvolutional {N}etworks.
\newblock In \emph{{P}roceedings of the {I}{E}{E}{E}/{C}{V}{F} conference on computer vision and {P}attern {R}ecognition {W}orkshops}, pp.\  76--77, 2020.

\bibitem[Badrinarayanan et~al.(2017)Badrinarayanan, Kendall, and Cipolla]{badrinarayanan2017segnet}
Badrinarayanan, V., Kendall, A., and Cipolla, R.
\newblock Segnet: A deep convolutional encoder–decoder architecture for image segmentation.
\newblock \emph{IEEE Transactions on Pattern Analysis and Machine Intelligence}, 39\penalty0 (12):\penalty0 2481--2495, 2017.
\newblock \doi{10.1109/TPAMI.2016.2644615}.

\bibitem[Benson et~al.(2024)Benson, Robin, Requena-Mesa, Alonso, Carvalhais, Cort{\'e}s, Gao, Linscheid, Weynants, and Reichstein]{benson2024multi}
Benson, V., Robin, C., Requena-Mesa, C., Alonso, L., Carvalhais, N., Cort{\'e}s, J., Gao, Z., Linscheid, N., Weynants, M., and Reichstein, M.
\newblock {M}ulti-modal learning for geospatial vegetation forecasting.
\newblock In \emph{{P}roceedings of the {I}{E}{E}{E}/{C}{V}{F} {C}onference on {C}omputer {V}ision and {P}attern {R}ecognition}, pp.\  27788--27799, 2024.

\bibitem[Chen et~al.(2018)Chen, Zhu, Papandreou, Schroff, and Adam]{chen2018encoder}
Chen, L.-C., Zhu, Y., Papandreou, G., Schroff, F., and Adam, H.
\newblock Encoder-decoder with atrous separable convolution for semantic image segmentation.
\newblock In \emph{ECCV}, 2018.

\bibitem[Chu et~al.(2019)Chu, Potetz, Wang, Howard, Song, Brucher, Leung, and Adam]{loc2}
Chu, G., Potetz, B., Wang, W., Howard, A., Song, Y., Brucher, F., Leung, T., and Adam, H.
\newblock {G}eo-aware networks for fine-grained recognition.
\newblock In \emph{{P}roceedings of the {I}{E}{E}{E}/{C}{V}{F} {I}nternational {C}onference on {C}omputer {V}ision {W}orkshops}, pp.\  0--0, 2019.

\bibitem[Clasen et~al.(2024)Clasen, Hackel, Burgert, Sumbul, Demir, and Markl]{bigearthnetv2}
Clasen, K.~N., Hackel, L.~W., Burgert, T., Sumbul, G., Demir, B., and Markl, V.
\newblock re{BEN}: {R}efined {B}ig{E}arth{N}et dataset for {R}emote {S}ensing {I}mage {A}nalysis.
\newblock \emph{{C}o{R}{R}}, abs/2407.03653, 2024.
\newblock URL \url{https://doi.org/10.48550/arXiv.2407.03653}.

\bibitem[Cong et~al.(2022)Cong, Khanna, Meng, Liu, Rozi, He, Burke, Lobell, and Ermon]{satmae}
Cong, Y., Khanna, S., Meng, C., Liu, P., Rozi, E., He, Y., Burke, M., Lobell, D., and Ermon, S.
\newblock {S}at{MAE}: {P}re-training transformers for temporal and multi-spectral satellite imagery.
\newblock \emph{{A}dvances in {N}eural {I}nformation {P}rocessing {S}ystems}, 35:\penalty0 197--211, 2022.

\bibitem[Darcet et~al.(2024)Darcet, Oquab, Mairal, and Bojanowski]{registers}
Darcet, T., Oquab, M., Mairal, J., and Bojanowski, P.
\newblock {V}ision {T}ransformers {N}eed {R}egisters.
\newblock In \emph{{T}he {T}welfth {I}nternational {C}onference on {L}earning {R}epresentations}, 2024.
\newblock URL \url{https://openreview.net/forum?id=2dnO3LLiJ1}.

\bibitem[Dosovitskiy et~al.(2021)Dosovitskiy, Beyer, Kolesnikov, Weissenborn, Zhai, Unterthiner, Dehghani, Minderer, Heigold, Gelly, Uszkoreit, and Houlsby]{vit}
Dosovitskiy, A., Beyer, L., Kolesnikov, A., Weissenborn, D., Zhai, X., Unterthiner, T., Dehghani, M., Minderer, M., Heigold, G., Gelly, S., Uszkoreit, J., and Houlsby, N.
\newblock An image is worth 16x16 words: Transformers for image recognition at scale.
\newblock In \emph{International Conference on Learning Representations (ICLR)}, 2021.
\newblock URL \url{https://arxiv.org/abs/2010.11929}.
\newblock arXiv:2010.11929.

\bibitem[Fan et~al.(2020)Fan, Wang, Cheng, and Tao]{fan2020ma}
Fan, T., Wang, X., Cheng, M., and Tao, D.
\newblock Ma-net: Multi-scale attention network for liver and tumor segmentation.
\newblock \emph{IEEE Access}, 8:\penalty0 179683--179691, 2020.
\newblock \doi{10.1109/ACCESS.2020.3025372}.

\bibitem[Fonte et~al.(2020)Fonte, Patriarca, Jesus, and Duarte]{osm2}
Fonte, C.~C., Patriarca, J., Jesus, I., and Duarte, D.
\newblock {A}utomatic extraction and filtering of openstreetmap data to generate training datasets for land use land cover classification.
\newblock \emph{{R}emote {S}ensing}, 12\penalty0 (20):\penalty0 3428, 2020.

\bibitem[Haklay \& Weber(2008)Haklay and Weber]{osm}
Haklay, M. and Weber, P.
\newblock {O}pen{S}treet{M}ap: {U}ser-generated street maps.
\newblock \emph{{I}{E}{E}{E} {P}ervasive computing}, 7\penalty0 (4):\penalty0 12--18, 2008.

\bibitem[He et~al.(2015)He, Zhang, Ren, and Sun]{resnet}
He, K., Zhang, X., Ren, S., and Sun, J.
\newblock {D}eep {R}esidual {L}earning for {I}mage {R}ecognition.
\newblock \emph{2016 {I}{E}{E}{E} {C}onference on {C}omputer {V}ision and {P}attern {R}ecognition ({C}{V}{P}{R})}, pp.\  770--778, 2015.
\newblock URL \url{https://api.semanticscholar.org/CorpusID:206594692}.

\bibitem[Hengl et~al.(2020)Hengl, Leal~Parente, Krizan, and Bonannella]{opentopo}
Hengl, T., Leal~Parente, L., Krizan, J., and Bonannella, C.
\newblock Continental europe digital terrain model at 30 m resolution based on gedi, icesat-2, aw3d, glo-30, eudem, merit dem and background layers.
\newblock \emph{Version Dataset v3. 0. Zenodo}, 2020.

\bibitem[Hou et~al.(2021)Hou, Liu, Zhang, and Li]{hou2021cunet}
Hou, Y., Liu, Z., Zhang, T., and Li, Y.
\newblock C-unet: Complement unet for remote sensing road extraction.
\newblock \emph{Sensors}, 21\penalty0 (6):\penalty0 2153, 2021.
\newblock \doi{10.3390/s21062153}.

\bibitem[Jia et~al.(2022)Jia, Tang, Chen, Cardie, Belongie, Hariharan, and Lim]{vpt}
Jia, M., Tang, L., Chen, B.-C., Cardie, C., Belongie, S., Hariharan, B., and Lim, S.-N.
\newblock {V}isual {P}rompt {T}uning.
\newblock In \emph{{E}uropean {C}onference on {C}omputer {V}ision ({E}{C}{C}{V})}, 2022.

\bibitem[Johnson et~al.(2022)Johnson, Treible, and Crispell]{opensentinelmap}
Johnson, N., Treible, W., and Crispell, D.
\newblock {O}pensentinelmap: {A} large-scale land use dataset using {O}pen{S}treet{M}ap and {S}entinel-2 imagery.
\newblock In \emph{{P}roceedings of the {I}{E}{E}{E}/{C}{V}{F} {C}onference on {C}omputer {V}ision and {P}attern {R}ecognition}, pp.\  1333--1341, 2022.

\bibitem[Klemmer et~al.(2025)Klemmer, Rolf, Robinson, Mackey, and Ru{\ss}wurm]{satclip}
Klemmer, K., Rolf, E., Robinson, C., Mackey, L., and Ru{\ss}wurm, M.
\newblock {S}at{CLIP}: {G}lobal, general-purpose location embeddings with satellite imagery.
\newblock In \emph{Proceedings of the AAAI conference on artificial intelligence}, 2025.

\bibitem[Long et~al.(2015)Long, Shelhamer, and Darrell]{long2015fully}
Long, J., Shelhamer, E., and Darrell, T.
\newblock Fully convolutional networks for semantic segmentation.
\newblock In \emph{Proceedings of the IEEE Conference on Computer Vision and Pattern Recognition}, pp.\  3431--3440, 2015.

\bibitem[Mac~Aodha et~al.(2019)Mac~Aodha, Cole, and Perona]{mac2019presence}
Mac~Aodha, O., Cole, E., and Perona, P.
\newblock {P}resence-only {G}eographical {P}riors for fine-grained image classification.
\newblock In \emph{{P}roceedings of the {I}{E}{E}{E}/{C}{V}{F} {I}nternational {C}onference on {C}omputer {V}ision}, pp.\  9596--9606, 2019.

\bibitem[Mai et~al.(2023)Mai, Lao, He, Song, and Ermon]{CSP}
Mai, G., Lao, N., He, Y., Song, J., and Ermon, S.
\newblock {C}sp: {S}elf-supervised contrastive spatial pre-training for geospatial-visual representations.
\newblock In \emph{{I}nternational {C}onference on {M}achine {L}earning}, pp.\  23498--23515. PMLR, 2023.

\bibitem[Nedungadi et~al.(2024)Nedungadi, Kariryaa, Oehmcke, Belongie, Igel, and Lang]{mmearth}
Nedungadi, V., Kariryaa, A., Oehmcke, S., Belongie, S., Igel, C., and Lang, N.
\newblock {M}{M}{E}arth: {E}xploring multi-modal pretext tasks for geospatial representation learning.
\newblock In \emph{European Conference on Computer Vision}, pp.\  164--182. Springer, 2024.

\bibitem[Patriarca et~al.(2019)Patriarca, Fonte, Estima, de~Almeida, and Cardoso]{osm3}
Patriarca, J., Fonte, C., Estima, J., de~Almeida, J.-P., and Cardoso, A.
\newblock {A}utomatic conversion of {O}{S}{M} data into {L}{U}{L}{C} maps: comparing {F}{O}{S}{S}4{G} based approaches towards an enhanced performance.
\newblock \emph{{O}pen {G}eospatial {D}ata, {S}oftware and {S}tandards}, 4:\penalty0 1--19, 2019.

\bibitem[Pickard et~al.(2015)Pickard, Daniel, Mehaffey, Jackson, and Neale]{enviroatlas}
Pickard, B.~R., Daniel, J., Mehaffey, M., Jackson, L.~E., and Neale, A.
\newblock {E}nviro{A}tlas: {A} new geospatial tool to foster ecosystem services science and resource management.
\newblock \emph{{E}cosystem {S}ervices}, 14:\penalty0 45--55, 2015.

\bibitem[Radford et~al.(2021)Radford, Kim, Hallacy, Ramesh, Goh, Agarwal, Sastry, Askell, Mishkin, Clark, et~al.]{clip}
Radford, A., Kim, J.~W., Hallacy, C., Ramesh, A., Goh, G., Agarwal, S., Sastry, G., Askell, A., Mishkin, P., Clark, J., et~al.
\newblock {L}earning transferable visual models from natural language supervision.
\newblock In \emph{{I}nternational conference on {M}achine {L}earning}, pp.\  8748--8763. PmLR, 2021.

\bibitem[Reed et~al.(2023)Reed, Gupta, Li, Brockman, Funk, Clipp, Keutzer, Candido, Uyttendaele, and Darrell]{scalemae}
Reed, C.~J., Gupta, R., Li, S., Brockman, S., Funk, C., Clipp, B., Keutzer, K., Candido, S., Uyttendaele, M., and Darrell, T.
\newblock {S}cale-mae: {A} scale-aware masked autoencoder for multiscale geospatial representation learning.
\newblock In \emph{{P}roceedings of the {I}{E}{E}{E}/{C}{V}{F} {I}nternational {C}onference on {C}omputer {V}ision}, pp.\  4088--4099, 2023.

\bibitem[Rolf et~al.(2021)Rolf, Proctor, Carleton, Bolliger, Shankar, Ishihara, Recht, and Hsiang]{mosaiks}
Rolf, E., Proctor, J., Carleton, T., Bolliger, I., Shankar, V., Ishihara, M., Recht, B., and Hsiang, S.
\newblock {A} generalizable and accessible approach to machine learning with global satellite imagery.
\newblock \emph{{N}ature communications}, 12\penalty0 (1):\penalty0 4392, 2021.

\bibitem[Rolf et~al.(2022)Rolf, Malkin, Graikos, Jojic, Robinson, and Jojic]{ipm}
Rolf, E., Malkin, N., Graikos, A., Jojic, A., Robinson, C., and Jojic, N.
\newblock Resolving label uncertainty with implicit posterior models.
\newblock In Cussens, J. and Zhang, K. (eds.), \emph{Proceedings of the Thirty-Eighth Conference on Uncertainty in Artificial Intelligence}, volume 180 of \emph{Proceedings of Machine Learning Research}, pp.\  1707--1717. PMLR, 01--05 Aug 2022.
\newblock URL \url{https://proceedings.mlr.press/v180/rolf22a.html}.

\bibitem[Rolf et~al.(2024)Rolf, Klemmer, Robinson, and Kerner]{rolfposition}
Rolf, E., Klemmer, K., Robinson, C., and Kerner, H.
\newblock {P}osition: {M}ission {C}ritical--{S}atellite {D}ata is a {D}istinct {M}odality in {M}achine {L}earning.
\newblock In \emph{{F}orty-first {I}nternational {C}onference on {M}achine {L}earning}, 2024.

\bibitem[Ronneberger et~al.(2015{\natexlab{a}})Ronneberger, Fischer, and Brox]{ronneberger2015u}
Ronneberger, O., Fischer, P., and Brox, T.
\newblock U-net: Convolutional networks for biomedical image segmentation.
\newblock In \emph{International Conference on Medical Image Computing and Computer-Assisted Intervention}, pp.\  234--241. Springer, 2015{\natexlab{a}}.

\bibitem[Ronneberger et~al.(2015{\natexlab{b}})Ronneberger, Fischer, and Brox]{unet}
Ronneberger, O., Fischer, P., and Brox, T.
\newblock {U}-net: {C}onvolutional networks for biomedical image segmentation.
\newblock In \emph{{M}edical image computing and computer-assisted intervention--{M}{I}{C}{C}{A}{I} 2015: 18th international conference, {M}unich, {G}ermany, {O}ctober 5-9, 2015, proceedings, part {I}{I}{I} 18}, pp.\  234--241. Springer, 2015{\natexlab{b}}.

\bibitem[Roscher et~al.(2024)Roscher, Russwurm, Gevaert, Kampffmeyer, Dos~Santos, Vakalopoulou, Hänsch, Hansen, Nogueira, Prexl, and Tuia]{datacentric}
Roscher, R., Russwurm, M., Gevaert, C., Kampffmeyer, M., Dos~Santos, J.~A., Vakalopoulou, M., Hänsch, R., Hansen, S., Nogueira, K., Prexl, J., and Tuia, D.
\newblock Better, not just more: Data-centric machine learning for earth observation.
\newblock \emph{IEEE Geoscience and Remote Sensing Magazine}, 12\penalty0 (4):\penalty0 335--355, 2024.
\newblock \doi{10.1109/MGRS.2024.3470986}.

\bibitem[Sosa et~al.(2025)Sosa, Rukhovich, Kacem, and Aouada]{multimae}
Sosa, J., Rukhovich, D., Kacem, A., and Aouada, D.
\newblock Multimae meets earth observation: Pre-training multi-modal multi-task masked autoencoders for earth observation tasks, 2025.
\newblock URL \url{https://arxiv.org/abs/2505.14951}.

\bibitem[Sumbul et~al.(2019)Sumbul, Charfuelan, Demir, and Markl]{bigearthnet}
Sumbul, G., Charfuelan, M., Demir, B., and Markl, V.
\newblock {B}igearthnet: {A} large-scale benchmark archive for remote sensing image understanding.
\newblock In \emph{{I}{G}{A}{R}{S}{S} 2019-2019 {I}{E}{E}{E} {I}nternational {G}eoscience and {R}emote {S}ensing {S}ymposium}, pp.\  5901--5904. IEEE, 2019.

\bibitem[Tang et~al.(2015)Tang, Paluri, Fei-Fei, Fergus, and Bourdev]{loc1}
Tang, K., Paluri, M., Fei-Fei, L., Fergus, R., and Bourdev, L.
\newblock {I}mproving image classification with location context.
\newblock In \emph{{P}roceedings of the {I}{E}{E}{E} {I}nternational {C}onference on {C}omputer {V}ision}, pp.\  1008--1016, 2015.

\bibitem[Touvron et~al.(2021)Touvron, Cord, Douze, Massa, Sablayrolles, and Jegou]{deit}
Touvron, H., Cord, M., Douze, M., Massa, F., Sablayrolles, A., and Jegou, H.
\newblock {T}raining data-efficient image transformers \& distillation through attention.
\newblock In Meila, M. and Zhang, T. (eds.), \emph{{P}roceedings of the 38th {I}nternational {C}onference on {M}achine {L}earning}, volume 139 of \emph{Proceedings of Machine Learning Research}, pp.\  10347--10357. PMLR, 18--24 Jul 2021.
\newblock URL \url{https://proceedings.mlr.press/v139/touvron21a.html}.

\bibitem[Vaswani et~al.(2017)Vaswani, Shazeer, Parmar, Uszkoreit, Jones, Gomez, and Łukasz Kaiser]{vaswani2017attention}
Vaswani, A., Shazeer, N., Parmar, N., Uszkoreit, J., Jones, L., Gomez, A., and Łukasz Kaiser.
\newblock {A}ttention is all you need.
\newblock \emph{{A}dvances in {N}eural {I}nformation {P}rocessing {S}ystems}, 2017.

\bibitem[Vivanco~Cepeda et~al.(2023)Vivanco~Cepeda, Nayak, and Shah]{geoclip}
Vivanco~Cepeda, V., Nayak, G.~K., and Shah, M.
\newblock {G}eoclip: {C}lip-inspired alignment between locations and images for effective worldwide geo-localization.
\newblock \emph{{A}dvances in {N}eural {I}nformation {P}rocessing {S}ystems}, 36:\penalty0 8690--8701, 2023.

\bibitem[Wang et~al.(2024)Wang, Cheng, Fang, Zhang, Duan, and Wang]{vpt2}
Wang, Y., Cheng, L., Fang, C., Zhang, D., Duan, M., and Wang, M.
\newblock {R}evisiting the {P}ower of {P}rompt for {V}isual {T}uning.
\newblock In \emph{{I}{C}{M}{L}}, 2024.
\newblock URL \url{https://openreview.net/forum?id=2Y93PtAqCl}.

\bibitem[Wang et~al.(2020)Wang, Yu, Wu, Zhang, Mao, Li, Feng, and Yin]{wang2020urban2vec}
Wang, Z., Yu, J., Wu, Z., Zhang, R., Mao, J., Li, L., Feng, Z., and Yin, J.
\newblock {U}rban2{V}ec: {I}ncorporating {S}treet {V}iew {I}magery and {P}{O}{I}{S} for {M}ulti-{M}odal {U}rban {N}eighborhood {E}mbedding.
\newblock In \emph{{P}roceedings of the 26th {A}{C}{M} {S}{I}{G}{K}{D}{D} {I}nternational {C}onference on {K}nowledge {D}iscovery \& {D}ata {M}ining}, pp.\  2068--2076. ACM, 2020.

\bibitem[Weng et~al.(2020)Weng, Xu, Xia, Zhang, Liu, and Xu]{weng2020water}
Weng, L., Xu, Y., Xia, M., Zhang, Y., Liu, J., and Xu, Y.
\newblock Water areas segmentation from remote sensing images using a separable residual segnet network.
\newblock \emph{ISPRS International Journal of Geo-Information}, 9\penalty0 (4):\penalty0 256, 2020.
\newblock \doi{10.3390/ijgi9040256}.

\bibitem[Xie et~al.(2021)Xie, Wang, Yu, Anandkumar, Alvarez, and Luo]{xie2021segformer}
Xie, E., Wang, W., Yu, Z., Anandkumar, A., Alvarez, J.~M., and Luo, P.
\newblock Segformer: Simple and efficient design for semantic segmentation with transformers.
\newblock In \emph{Advances in Neural Information Processing Systems}, volume~34, pp.\  12077--12090, 2021.

\bibitem[Yeh et~al.(2021)Yeh, Meng, Wang, Driscoll, Rozi, Liu, Lee, Burke, Lobell, and Ermon]{sustainbench}
Yeh, C., Meng, C., Wang, S., Driscoll, A., Rozi, E., Liu, P., Lee, J., Burke, M., Lobell, D.~B., and Ermon, S.
\newblock {S}ustain{B}ench: {B}enchmarks for {M}onitoring the {S}ustainable {D}evelopment {G}oals with {M}achine {L}earning.
\newblock In \emph{{T}hirty-fifth {C}onference on {N}eural {I}nformation {P}rocessing {S}ystems {D}atasets and {B}enchmarks {T}rack (Round 2)}, 2021.
\newblock URL \url{https://openreview.net/forum?id=5HR3vCylqD}.

\bibitem[Yin et~al.(2019)Yin, Liu, Zhang, Wang, Shah, and Zimmermann]{gps2vec}
Yin, Y., Liu, Z., Zhang, Y., Wang, S., Shah, R.~R., and Zimmermann, R.
\newblock {GPS}2{V}ec: {T}owards generating worldwide {GPS} embeddings.
\newblock In \emph{{P}roceedings of the 27th {A}{C}{M} {S}{I}{G}{S}{P}{A}{T}{I}{A}{L} {I}nternational {C}onference on {A}dvances in {G}eographic {I}nformation {S}ystems}, pp.\  416--419, 2019.

\bibitem[Yuan et~al.(2022)Yuan, Wang, and Xu]{yuan2022shift}
Yuan, W., Wang, J., and Xu, W.
\newblock Shift pooling pspnet: Rethinking pspnet for building extraction in remote sensing images from entire local feature pooling.
\newblock \emph{Remote Sensing}, 14\penalty0 (19):\penalty0 4889, 2022.
\newblock \doi{10.3390/rs14194889}.

\bibitem[Zanaga et~al.(2022)Zanaga, Van De~Kerchove, Daems, De~Keersmaecker, Brockmann, Kirches, Wevers, Cartus, Santoro, Fritz, et~al.]{zanaga2022esa}
Zanaga, D., Van De~Kerchove, R., Daems, D., De~Keersmaecker, W., Brockmann, C., Kirches, G., Wevers, J., Cartus, O., Santoro, M., Fritz, S., et~al.
\newblock {E}{S}{A} {W}orld{C}over 10 m 2021 v200.
\newblock 2022.

\bibitem[Zhao et~al.(2017)Zhao, Shi, Qi, Wang, and Jia]{zhao2017pyramid}
Zhao, H., Shi, J., Qi, X., Wang, X., and Jia, J.
\newblock Pyramid scene parsing network.
\newblock In \emph{Proceedings of the IEEE Conference on Computer Vision and Pattern Recognition}, pp.\  2881--2890, 2017.

\end{thebibliography}
\bibliographystyle{icml2025}

%%%%%%%%%%%%%%%%%%%%%%%%%%%%%%%%%%%%%%%%%%%%%%%%%%%%%%%%%%%%%%%%%%%%%%%%%%%%%%%
%%%%%%%%%%%%%%%%%%%%%%%%%%%%%%%%%%%%%%%%%%%%%%%%%%%%%%%%%%%%%%%%%%%%%%%%%%%%%%%
% APPENDIX
%%%%%%%%%%%%%%%%%%%%%%%%%%%%%%%%%%%%%%%%%%%%%%%%%%%%%%%%%%%%%%%%%%%%%%%%%%%%%%%
%%%%%%%%%%%%%%%%%%%%%%%%%%%%%%%%%%%%%%%%%%%%%%%%%%%%%%%%%%%%%%%%%%%%%%%%%%%%%%%
\newpage
\appendix
\onecolumn
\clearpage
\setcounter{page}{1}

\clearpage
\onecolumn
\setcounter{page}{1}

\section{Experimental Setup}
\subsection{FCN on EnviroAtlas Land Cover Segmentation with \texttt{STACK}, \texttt{PROC-STACK}: } 

We train on EnviroAtlas' train images in Pittsburgh, PA on a 5-layer Fully Convolutional Network with 64 filters and an output smoothing of $10^{-2}$. A batch size of $128$ and a learning rate of $1e-3$ are fixed across all training data subsets and random seeds reported in \Cref{fig:enviroatlas}. We fix the lower bound learning rate to $1e-7$. Table \ref{tab:subset_epochs} reports the number of training epochs each data-efficient FCN is trained on. Note that FCNs trained on $1\%$ of EnviroAtlas' training data for 700 epochs trigger our early-stopping logic between epoch 200-300. We use TorchGeo's \verb|RandomGeoSampler| with an input image size of 128. Our test dataset uses TorchGeo's \verb|GridGeoSampler| with an input image size of $256$ and a stride of $512$ to avoid overlapping image patches. Our multi-modal inputs include a road, water, waterway, and waterbody footprint from \citep{osm}. 

\paragraph{Hand‐crafted prior generation process.}
\label{sec:priorgen}
In our PROC‐STACK experiments, the hand‐crafted prior \(f(x_i)\equiv p_i(\ell)\) is constructed exactly as in \cite{ipm} (“Coarse data in weakly supervised segmentation”, §3), using the NLCD 30 m land‐cover map to induce per‐pixel beliefs over our four high‐resolution classes. Concretely, we first compute the empirical co‐occurrence matrix
\[
P(\ell \mid c)
\;=\;
\frac{\bigl|\{\,\text{high‐res label}=\ell,\;\text{NLCD class}=c\}\bigr|}
{\sum_{\ell'}\bigl|\{\,\text{high‐res label}=\ell',\;\text{NLCD class}=c\}\bigr|}
\]
from a held‐out set of aligned NAIP+NLCD+Land Cover tiles. Then, for each pixel \(i\) with NLCD class \(c_i\), we set
\[
p_i(\ell)\;=\;P(\ell\mid c_i)
\]
and apply a small Gaussian blur (\(\sigma=1\) pixel) to smooth block artifacts. In PROC‐STACK mode, we further enrich this prior with binary auxiliary masks (roads, buildings, waterways): for each feature \(j\), we define
\[
M_j(i)=
\begin{cases}
1, & \text{if feature }j\text{ lies within a 10 m radius of pixel }i,\\
0, & \text{otherwise,}
\end{cases}
\]
and boost the corresponding class by adding a fixed weight \(w_j\) to \(p_i(\ell=j)\). Finally, we re-normalize \(p_i(\ell)\) so that \(\sum_{\ell}p_i(\ell)=1\). This yields a spatially varying, hand‐crafted prior that both encodes coarse NLCD statistics and injects domain knowledge via auxiliary GIS layers, as required by the PROC‐STACK formulation.

\begin{table}[h!]
  \centering
  \setlength{\tabcolsep}{5pt} % reduce column spacing
  \renewcommand{\arraystretch}{1.0} % reduce row spacing
  \begin{tabular}{cc}
    \toprule
    \textbf{Subset Size} & \textbf{Training Epochs} \\
    \midrule
    100\% & 7 \\
    75\%  & 9 \\
    50\%  & 14 \\
    35\%  & 20 \\
    20\%  & 35 \\
    10\%  & 70 \\
    5\%   & 140 \\
    2\%   & 350 \\
    1\%   & 700 \\
    \bottomrule
  \end{tabular}
  \caption{Training epochs scaled by subset size for all label-efficiency experiments.}
  \label{tab:subset_epochs}
\end{table}

\subsection{ViT on BigEarthNetv2.0 Multi-label classification with \texttt{TOKEN-FUSE}}

\begin{figure*}[t!]
    \centering
    \includegraphics[width=\linewidth]{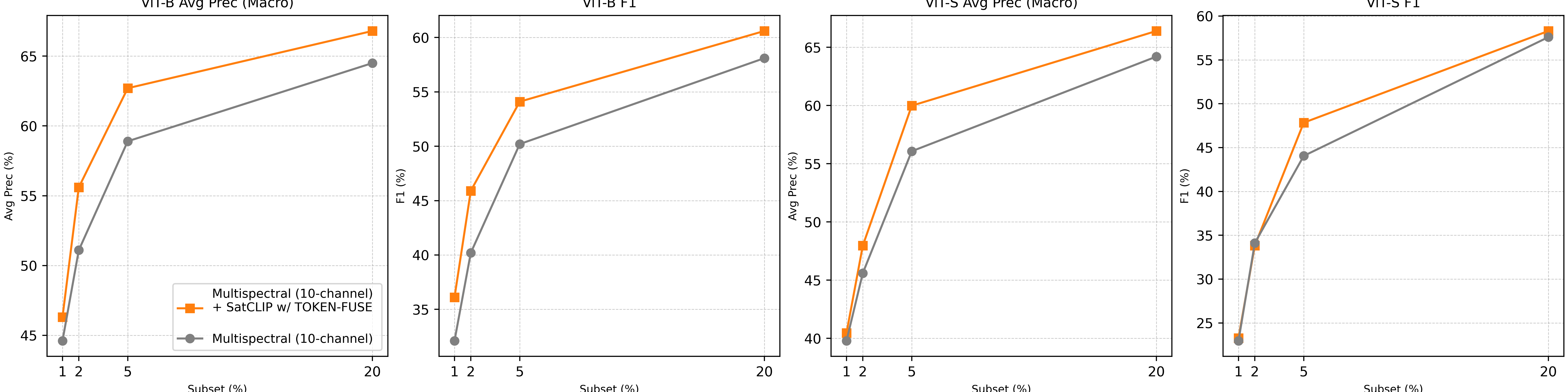}
    \caption{\textbf{Label efficiency of a ViT trained with an auxiliary SatCLIP token. } \textbf{Left: } ViT-Base (86M trainable parameters). SatCLIP linear projection layer mapped to embedding dimension of $768$. \textbf{Right: } ViT-small (22M trainable params), SatCLIP linear projection layer mapped to embedding dimension of $384$.}
    \label{fig:vit_ben_plots}
\end{figure*}

Our experiments with the Vision Transformer (ViT) use a ViT-Base and a ViT-Small ($86$M and $22$M trainable parameters) with a fixed patch size of $8$. All ViTs are randomly initialized for a fixed random seed. We prepend a learnable location token \(x_{\text{loc}} \in \mathbb{R}^D\) to the input sequence in addition to a class token \(x_{\text{cls}} \in \mathbb{R}^D\) and \(N\) patch tokens \(x_{\text{patch}}^{(i)} \in \mathbb{R}^D\). The token sequence is given by
\[
X_{\text{tokens}} = \Bigl[ x_{\text{cls}}; \; x_{\text{loc}}; \; x_{\text{patch}}^{(1)}, \dots, x_{\text{patch}}^{(N)} \Bigr] \in \mathbb{R}^{(N+2)\times D},
\]
We add corresponding learnable positional embeddings 
\[
E_{\text{pos}} = \Bigl[ e_{\text{cls}}; \; e_{\text{loc}}; \; e_{\text{patch}}^{(1)}, \dots, e_{\text{patch}}^{(N)} \Bigr].
\]
Our final sequence is \(z_0 = X_{\text{tokens}} + E_{\text{pos}}\). With the addition of the auxiliary SatCLIP token with \texttt{TOKEN-FUSE}, our sequence length is increased by one and allows the model to jointly encode class and location information.
\paragraph{Why SatCLIP?} SatCLIP is currently the only location encoder in previous work that is pre-trained on Sentinel-2 satellite imagery, hence making it a suitable candidate for our experiments that primarily train, validate, and test on geospatial satellite imagery. Future work will incorporate the label-efficiency and out-of-sample performance for SatML models trained with newer location encoders that are pre-trained with satellite or geospatial imagery. 

Our experiments on the BigEarthNetv2.0 dataset use a batch size of $700$ and run for $15$ epochs ($5$ warmup epochs) at a base learning rate of $5e-4$. We use a dropout rate of $0.15$ to prevent overfitting across all settings (Finetuned SatCLIP (\texttt{FT}), Frozen SatCLIP (\texttt{F}), and Register token). We record macro and micro-averaged average precision, recall, and F1 score in addition to class-wise accuracies. \Cref{fig:vit_ben_plots} shows label-efficiency results (similar to \Cref{tab:vit_table}) of a frozen SatCLIP auxiliary token with a learnable linear projection layer on the BigEarthNet2.0 dataset. 

\subsection{U-Net on SustainBench Field Boundary Delineation with \texttt{STACK}}
Our standard U-Net setup consists of 4 downsampling blocks, a bottleneck, and corresponding upsampling blocks with skip connections. Input images are georeferenced with a pre-processed OSM raster and are stored as an HDF5 dataset with 7 total channels. We use a random crop, horizontal, and vertical flip augmentation during training and a center crop for evaluation. The model is trained for 20 epochs with a batch size of 48 at a learning rate of \(1\times10^{-4}\). A learning rate scheduler cognizant of validation loss plateaus is used (factor 0.5, patience 5). We record the Dice coefficient, and the IoU score.

\paragraph{Ablations with model architectures: } We conduct a broad survey of commonly used SatML model architectures for semantic segmentation tasks from published work spanning 2020 to 2025. We find that most commonly used segmentation architectures include:
\begin{itemize}
  \item Fully Convolutional Networks (FCN) \cite{long2015fully}
  \item U-Net \cite{ronneberger2015u, hou2021cunet}
  \item SegNet \cite{badrinarayanan2017segnet, weng2020water}
  \item PSPNet \cite{zhao2017pyramid, yuan2022shift}
  \item DeepLabv3+ \cite{chen2018encoder}
  \item SegFormer \cite{xie2021segformer}
  \item MA-Net \cite{fan2020ma}
\end{itemize}

We choose 4 commonly used segmentation model architectures from the list above, and perform the label-efficiency experiments similar to \Cref{sec:results-data-efficiency} with and without an auxiliary geographic input of an OSM and EU-DEM raster layer. From \Cref{tab:susbench_model_ablations}, we see that our performance improvements hold consistently over all data subsets with the auxiliary geographic input.

\subsection{ResNet50 on USAVars Regression with \texttt{STACK}}
Our generated USAVars dataset comprises images with 7 channels and corresponding scalar labels. A custom \texttt{HDF5Dataset} class is used to load the data. For 7-channel inputs, the first four channels are normalized to \([0,1]\) by division by 255, while channels 4--6 are scaled from the original categorical values returned from the OSM API to the RGB space. Random cropping (to an image size of 256), horizontal, and vertical flips are applied during training, while a center crop is used for validation and testing. To accommodate inputs with 4 or 7 channels, the initial convolutional layer of ResNet50 is re-initialized accordingly. The final fully-connected layer is replaced with a linear layer outputting a single value for regression. We use a base learning rate of $1e-4$ with a batch size of $512$. We train the model for $20$ epochs. All experiments are seeded for reproducibility and results are reported over five random seeds. We record the mean squared error loss and the \(R^2\) score.

\begin{figure*}[ht!]
    \centering
    \includegraphics[width=\linewidth]{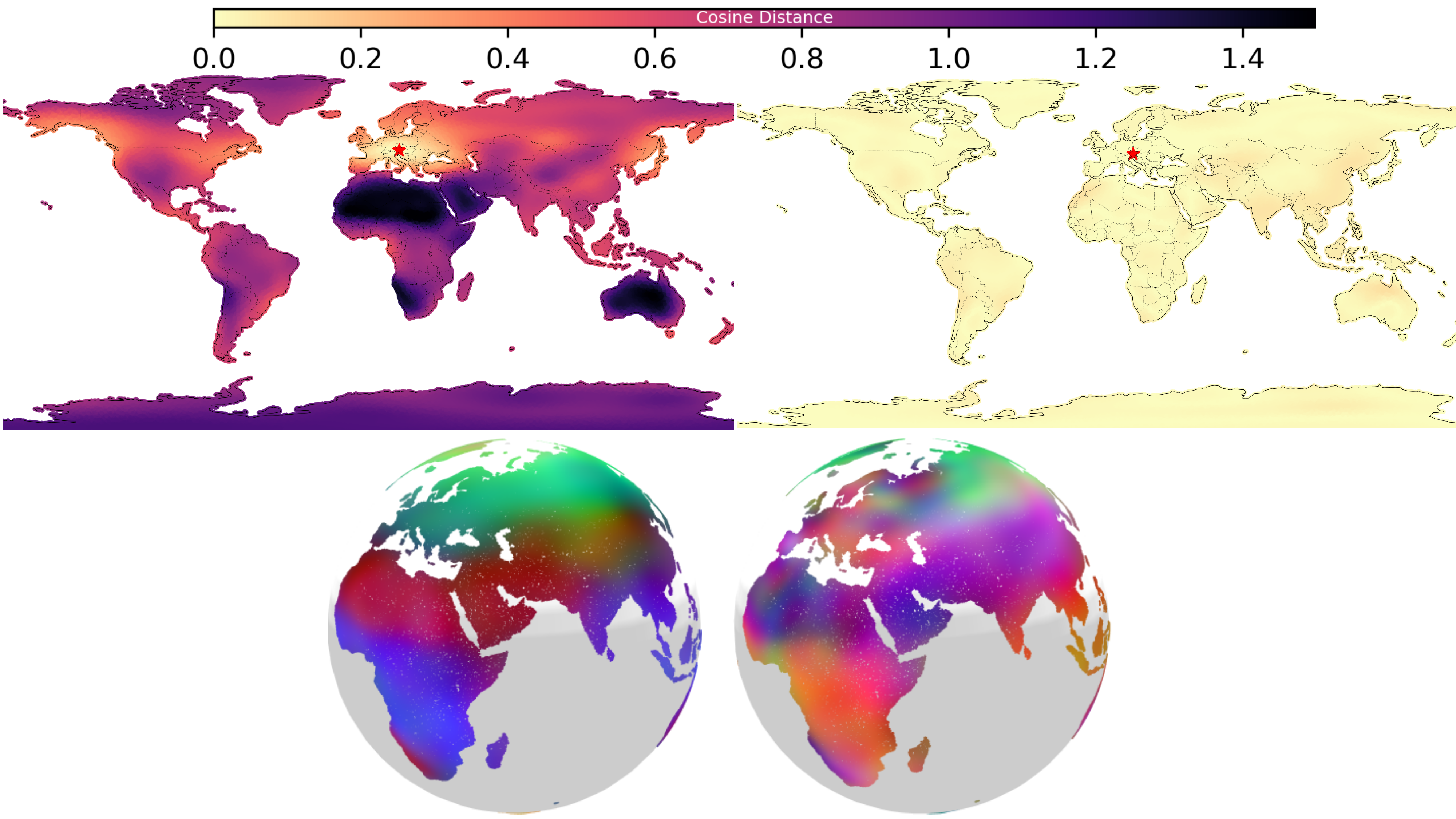}
    \caption{\textbf{Qualitative Result: Frozen \texttt{F} vs Finetuned \texttt{FT} SatCLIP auxiliary token} [Top-left] Cosine distance of standard SatCLIP embeddings to a fixed reference point in Austria. [Top-Right] Absolute difference between cosine distances between our \texttt{F} SatCLIP location encoder + trained linear projection layer and original SatCLIP location encoder cosine distances. [Bottom] Global PCA embeddings of \texttt{F} vs \texttt{FT} SatCLIP auxiliary token with \texttt{TOKEN-FUSE}}
    \label{fig:satclip_supplementary}
\end{figure*}

\begin{figure}[b!]
    \centering
    \includegraphics[width=\textwidth]{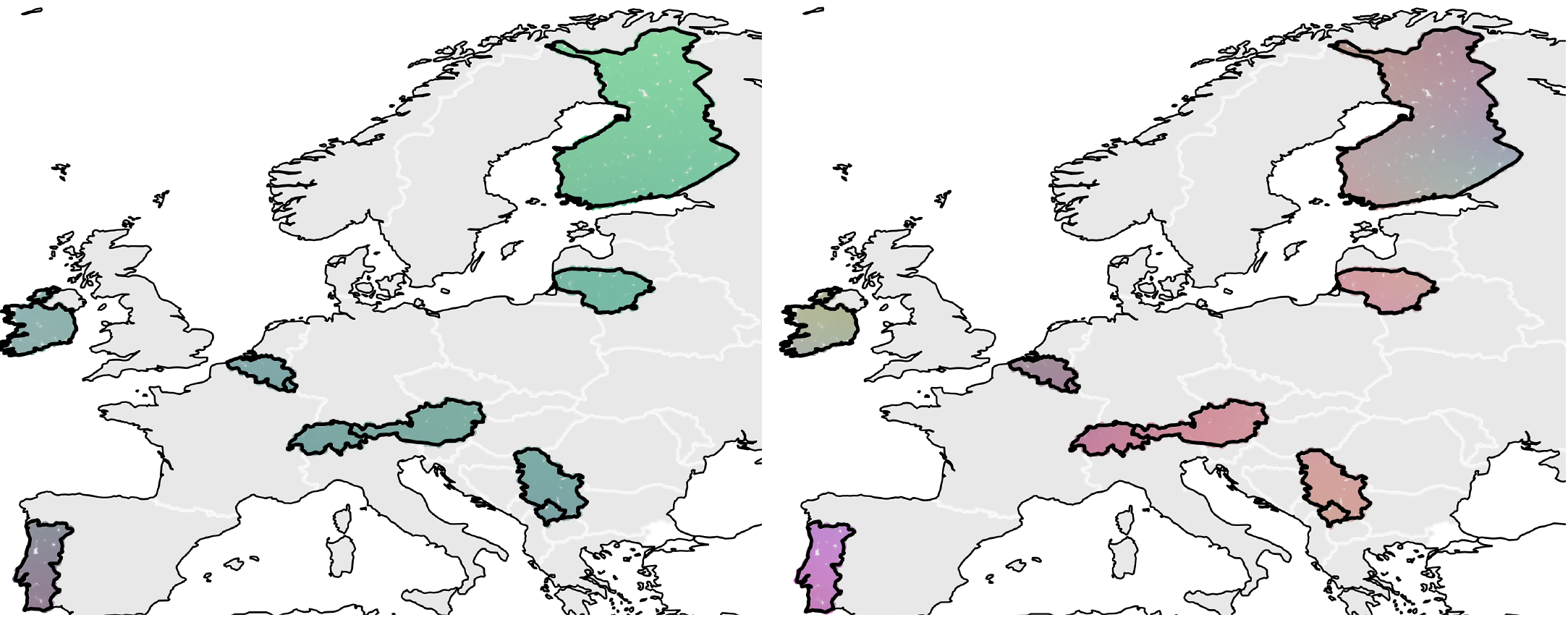}%
    \quad
    \caption{\textbf{Qualitative result: Frozen vs Finetuned SatCLIP auxiliary ViT token on the BigEarthNetv2.0 land-cover classification task:}  
      Maps: PCA embeddings of the SatCLIP tokens: frozen (left) vs finetuned (right) on 10 European countries covered by the BigEarthNetv2.0 dataset.}
    \label{fig:ft_satclip}
\end{figure}

\clearpage
\section{Qualitative Result: \texttt{TOKEN-FUSE}}
To qualitatively evaluate the quality of embeddings learned by our linear projection layer, which is responsible for mapping the 256-dimensional SatCLIP embeddings to the token size expected by the ViT, we calculate the disagreement of this learned layer with a standard SatCLIP location encoder. With a fixed reference SatCLIP embedding in Austria ($E_{\text{Austria}}$), we calculate the cosine distance between SatCLIP embeddings of 200,000 global, randomly sampled SatCLIP embeddings with $E_{\text{Austria}}$. The disagreement of our learned linear projection layer is calculated by repeating the same procedure after passing standard SatCLIP embeddings through the learned linear projection layer before calculating the cosine distance. \Cref{fig:satclip_supplementary} [top-right] shows that our learned linear projection layer successfully maps SatCLIP embeddings to the SatCLIP auxiliary token without a significant disagreement from original embeddings. \Cref{fig:satclip_supplementary}[Bottom] also shows a PCA visualization of a frozen (\texttt{F}) vs finetuned (\texttt{FT}) SatCLIP auxiliary token's embeddings mapped to RGB space. \Cref{fig:ft_satclip} surprisingly shows these PCA embeddings for countries covered by the train-split of the BigEarthNetv2.0 dataset \citep{bigearthnetv2}. These results support our observation in \Cref{fig:ft_satclip} that show that a finetuned SatCLIP token with \texttt{TOKEN-FUSE} learns high-resolution, arbitrary information compared to a frozen token. 

\begin{figure}[ht!]
    \centering
    \includegraphics[width=0.55\linewidth]{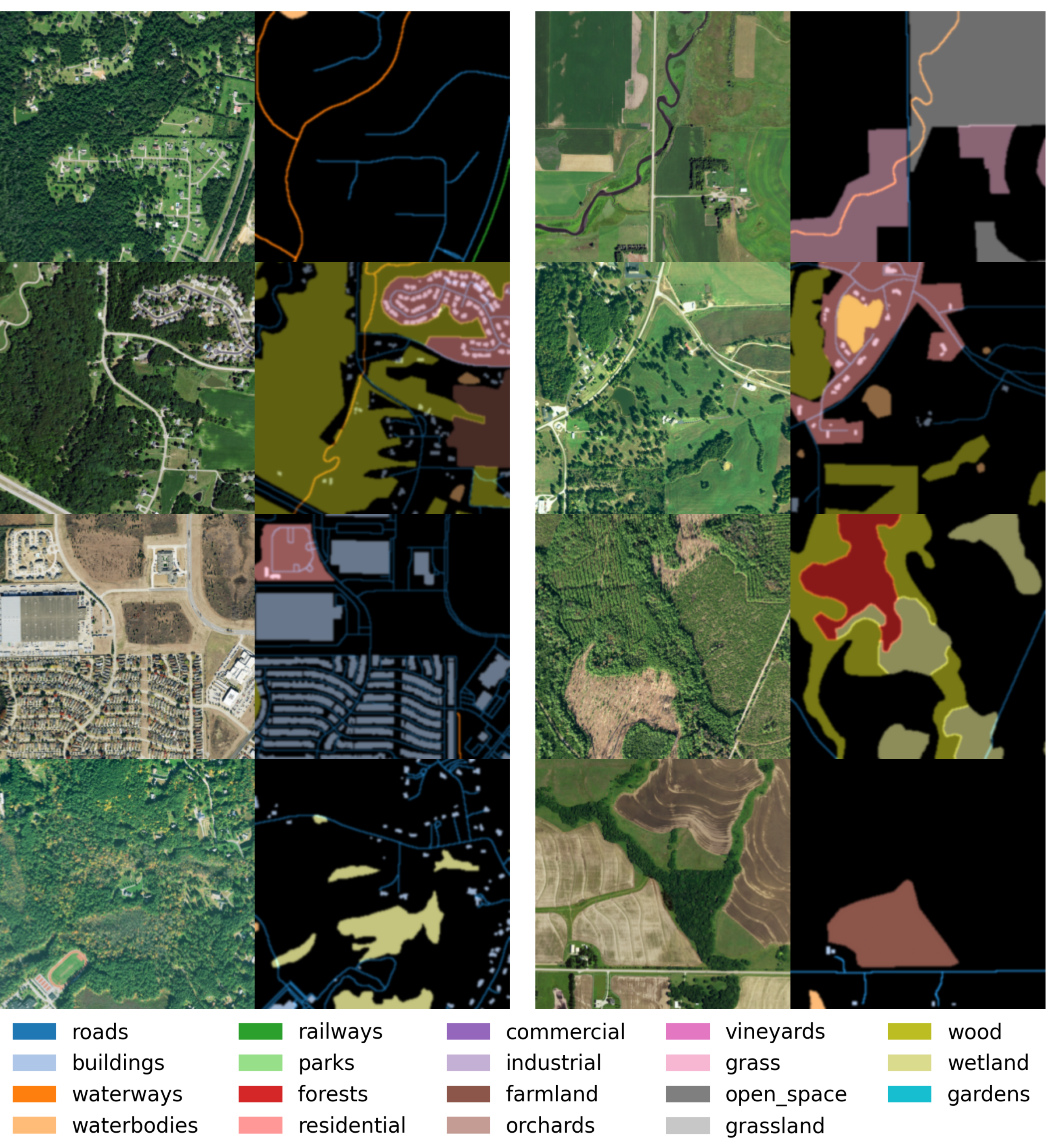}
    \caption{\textbf{NAIP Imagery from USAVars \citep{mosaiks} georeferenced with our OpenStreetMaps (OSM) raster geographic data-layer: } OSM products are smoothed with a Gaussian Kernel and pre-processed to RGB space.}
    \label{fig:sample_osm}
\end{figure}
\begin{table*}[ht]
  \centering
  {\normalsize
    \caption{\textbf{Performance and Label Efficiency of commonly used SatML semantic segmentation model architectures on the SustainBench field boundary delineation dataset with and without an OSM and EU-DEM auxiliary geographic data layer:} Model choices are informed by a surveying the SatML segmentation model literature spanning five years. Test Dice and IoU scores reported based on a hold-out validation set. Results averaged over 3 random seeds. Bolded numbers indicate best model performance for a given data subset.}
    \label{tab:susbench_model_ablations}
  }
  \scriptsize
  \resizebox{\textwidth}{!}{%
    \begin{tabular}{clcccc}
      \toprule
      \multirow{2}{*}{\textbf{Subset}} & \multirow{2}{*}{\textbf{Model}} &
        \multicolumn{2}{c}{\textbf{RGB}} &
        \multicolumn{2}{c}{\textbf{RGB $+$ OSM $+$ EU-DEM w/ \texttt{STACK}}} \\
      \cmidrule(lr){3-4} \cmidrule(lr){5-6}
      & & IoU & Dice & IoU & Dice \\
      \midrule

      \multirow{4}{*}{0.01}
        & \texttt{deeplabv3+} & 0.163\,$\pm$0.046 & 0.261\,$\pm$0.079
                             & 0.180\,$\pm$0.051 & 0.287\,$\pm$0.072 \\
        & \texttt{fpn}        & 0.133\,$\pm$0.040 & 0.222\,$\pm$0.063
                             & 0.135\,$\pm$0.042 & 0.224\,$\pm$0.065 \\
        & \texttt{pspnet}     & 0.125\,$\pm$0.043 & 0.207\,$\pm$0.074
                             & 0.132\,$\pm$0.048 & 0.221\,$\pm$0.066 \\
        & \texttt{unetpp}     & 0.251\,$\pm$0.006 & 0.397\,$\pm$0.008
                             & 0.259\,$\pm$0.007 & 0.407\,$\pm$0.009 \\
      \addlinespace

      \multirow{4}{*}{0.05}
        & \texttt{deeplabv3+} & 0.293\,$\pm$0.009 & 0.445\,$\pm$0.012
                             & \textbf{0.311\,$\pm$0.006} & 0.465\,$\pm$0.008 \\
        & \texttt{fpn}        & 0.114\,$\pm$0.024 & 0.197\,$\pm$0.036
                             & 0.120\,$\pm$0.021 & 0.207\,$\pm$0.032 \\
        & \texttt{pspnet}     & 0.156\,$\pm$0.025 & 0.262\,$\pm$0.038
                             & 0.148\,$\pm$0.036 & 0.246\,$\pm$0.054 \\
        & \texttt{unetpp}     & 0.318\,$\pm$0.007 & 0.475\,$\pm$0.008
                             & 0.333\,$\pm$0.008 & 0.491\,$\pm$0.009 \\
      \addlinespace

      \multirow{4}{*}{0.10}
        & \texttt{deeplabv3+} & 0.317\,$\pm$0.004 & 0.472\,$\pm$0.005
                             & \textbf{0.333\,$\pm$0.007} & \textbf{0.490\,$\pm$0.009} \\
        & \texttt{fpn}        & 0.123\,$\pm$0.012 & 0.212\,$\pm$0.019
                             & 0.139\,$\pm$0.010 & 0.238\,$\pm$0.015 \\
        & \texttt{pspnet}     & 0.157\,$\pm$0.007 & 0.264\,$\pm$0.011
                             & 0.169\,$\pm$0.016 & 0.281\,$\pm$0.025 \\
        & \texttt{unetpp}     & 0.363\,$\pm$0.004 & 0.524\,$\pm$0.004
                             & \textbf{0.377\,$\pm$0.004} & \textbf{0.538\,$\pm$0.004} \\
      \addlinespace

      \multirow{4}{*}{0.20}
        & \texttt{deeplabv3+} & 0.326\,$\pm$0.003 & 0.482\,$\pm$0.003
                             & \textbf{0.343\,$\pm$0.004} & \textbf{0.500\,$\pm$0.004} \\
        & \texttt{fpn}        & 0.193\,$\pm$0.008 & 0.314\,$\pm$0.011
                             & \textbf{0.213\,$\pm$0.004} & \textbf{0.342\,$\pm$0.006} \\
        & \texttt{pspnet}     & 0.153\,$\pm$0.002 & 0.258\,$\pm$0.005
                             & 0.154\,$\pm$0.002 & 0.258\,$\pm$0.003 \\
        & \texttt{unetpp}     & 0.385\,$\pm$0.003 & 0.548\,$\pm$0.004
                             & \textbf{0.405\,$\pm$0.002} & \textbf{0.567\,$\pm$0.002} \\
      \addlinespace

      \multirow{4}{*}{0.35}
        & \texttt{deeplabv3+} & 0.343\,$\pm$0.004 & 0.501\,$\pm$0.004
                             & \textbf{0.360\,$\pm$0.003} & \textbf{0.520\,$\pm$0.003} \\
        & \texttt{fpn}        & 0.241\,$\pm$0.010 & 0.377\,$\pm$0.012
                             & \textbf{0.266\,$\pm$0.011} & \textbf{0.409\,$\pm$0.014} \\
        & \texttt{pspnet}     & 0.163\,$\pm$0.006 & 0.272\,$\pm$0.009
                             & 0.164\,$\pm$0.004 & 0.273\,$\pm$0.005 \\
        & \texttt{unetpp}     & 0.391\,$\pm$0.002 & 0.554\,$\pm$0.002
                             & \textbf{0.415\,$\pm$0.003} & \textbf{0.578\,$\pm$0.003} \\
      \addlinespace

      \multirow{4}{*}{0.50}
        & \texttt{deeplabv3+} & 0.353\,$\pm$0.003 & 0.513\,$\pm$0.003
                             & \textbf{0.363\,$\pm$0.003} & \textbf{0.522\,$\pm$0.003} \\
        & \texttt{fpn}        & 0.253\,$\pm$0.005 & 0.393\,$\pm$0.007
                             & \textbf{0.285\,$\pm$0.004} & \textbf{0.433\,$\pm$0.005} \\
        & \texttt{pspnet}     & 0.167\,$\pm$0.008 & 0.278\,$\pm$0.011
                             & 0.174\,$\pm$0.002 & 0.289\,$\pm$0.002 \\
        & \texttt{unetpp}     & 0.398\,$\pm$0.001 & 0.561\,$\pm$0.001
                             & \textbf{0.420\,$\pm$0.002} & \textbf{0.582\,$\pm$0.002} \\
      \addlinespace

      \multirow{4}{*}{0.75}
        & \texttt{deeplabv3+} & 0.358\,$\pm$0.001 & 0.518\,$\pm$0.001
                             & \textbf{0.383\,$\pm$0.002} & \textbf{0.545\,$\pm$0.002} \\
        & \texttt{fpn}        & 0.285\,$\pm$0.007 & 0.433\,$\pm$0.008
                             & \textbf{0.315\,$\pm$0.004} & \textbf{0.468\,$\pm$0.005} \\
        & \texttt{pspnet}     & 0.187\,$\pm$0.007 & 0.306\,$\pm$0.010
                             & 0.194\,$\pm$0.007 & 0.317\,$\pm$0.010 \\
        & \texttt{unetpp}     & 0.402\,$\pm$0.002 & 0.564\,$\pm$0.002
                             & \textbf{0.430\,$\pm$0.001} & \textbf{0.593\,$\pm$0.001} \\
      \addlinespace

      \multirow{4}{*}{1.00}
        & \texttt{deeplabv3+} & 0.368\,$\pm$0.003 & 0.529\,$\pm$0.003
                             & \textbf{0.390\,$\pm$0.004} & \textbf{0.553\,$\pm$0.004} \\
        & \texttt{fpn}        & 0.311\,$\pm$0.003 & 0.464\,$\pm$0.003
                             & \textbf{0.335\,$\pm$0.004} & \textbf{0.491\,$\pm$0.004} \\
        & \texttt{pspnet}     & 0.199\,$\pm$0.005 & 0.323\,$\pm$0.007
                             & 0.204\,$\pm$0.003 & 0.330\,$\pm$0.004 \\
        & \texttt{unetpp}     & 0.408\,$\pm$0.002 & 0.571\,$\pm$0.002
                             & \textbf{0.436\,$\pm$0.001} & \textbf{0.598\,$\pm$0.001} \\
      \bottomrule
    \end{tabular}%
  }
  \normalsize
\end{table*}

\end{document}